\documentclass[]{M-PSI}

\usepackage[linesnumbered,ruled,vlined]{algorithm2e}
\usepackage{algpseudocode}
\usepackage{multirow}
\usepackage{graphicx}
\usepackage{dsfont}
\usepackage{subcaption} 
\usepackage{amsfonts} 
\usepackage{pifont}

\usepackage[most]{tcolorbox}
\usepackage{enumitem}
\definecolor{myGreen}{RGB}{68, 166, 142}
\tcbuselibrary{breakable}
\definecolor{lightpink}{RGB}{255, 228, 240}
\definecolor{lightgreen}{RGB}{230, 250, 230}
\definecolor{lightblue}{RGB}{230, 240, 255}
\definecolor{lightorange}{RGB}{255, 240, 220}
\definecolor{lavender}{RGB}{240, 230, 250}
\SetKwIF{If}{ElseIf}{Else}{if}{then}{else if}{else}{} 
\SetKwFor{ForEach}{foreach}{do}{}
\usepackage{longtable}
\usepackage{booktabs}

\title{Multi-Resident Data Association in Smart Homes via MDP Modeling and Behavior Cloning}

\author{
\AuthorEntry{Xi Chen}{,}{1,2}{}{0009-0004-0052-571X}
\AuthorEntry{Julien Cumin}{,}{2}{}{0000-0001-8846-5547}
\AuthorEntry{Fano Ramparany}{,}{2}{}{0000-0002-0617-7769}
\AuthorEntry{Dominique Vaufreydaz}{}{1}{https://research.vaufreydaz.org/}{0000-0002-8825-0973}\\
\vspace{0.5em}
\textnormal{\normalsize{
{$^1$ Univ. Grenoble Alpes, CNRS, Grenoble INP, LIG, 38000 Grenoble, France}\\ %
{$^2$ Orange Research}
}}
} %

\begin{document}

\vspace{1cm}
\begin{abstract}
\vspace{0.5em}
Ambient sensors offer a privacy-preserving solution for activity recognition in smart homes, yet they face a critical challenge in multi-resident environments known as Data Association, which aims to identify the resident triggering a specific anonymous sensor event. Existing approaches fundamentally fail to maintain global consistency by often yielding conflicting associations for silent residents. Additionally, these methods frequently suffer from severe trajectory fragmentation caused by the ontological mismatch of applying continuous physical motion models to discrete logical behaviors. To address these limitations, we propose a novel framework that reformulates data association as a sequential decision-making task within a Markov Decision Process (MDP). We solve the defined MDP via Behavior Cloning (BC) to avoid complex reward engineering by directly learning from expert demonstrations. We instantiate the policy using two complementary architectures, including a lightweight Transformer that treats association as a multi-label classification task over linearized event histories, and a fine-tuned Large Language Model (LLM) that reformulates the problem as conditional text generation to leverage pre-trained semantic knowledge. Extensive evaluations on three diverse datasets demonstrate that our approach significantly outperforms state-of-the-art baselines, particularly in resolving concurrent multi-resident events. Furthermore, the LLM-based policy shows preliminary cross-environment transfer potential under benchmark settings, while its robustness in real deployments still requires further validation.
\end{abstract}


\section{Introduction}
\label{sect:introduction}
Ambient sensors serve as a practical foundation for intelligent smart-home services because they are low-power, unobtrusive, and privacy-preserving, making them particularly attractive for daily residential monitoring compared to more intrusive modalities such as cameras or wearables. These properties have fueled a wave of sophisticated research into high-level behavioral analytics, including health monitoring, home safety, and building automation~\cite{acampora2013survey, riboni2016smartfaber, rahim2023enhancing, chan2024home, ahmadi2018real, lashkari2019energy}. However, a fundamental mismatch emerges when these applications move beyond controlled laboratory studies: most experimental settings simplify the problem to single-resident households, whereas real-world homes are commonly shared by multiple occupants. Under these realistic multi-resident conditions, the identity-blind nature of ambient observations becomes a fundamental and shared barrier. In all of these settings, the practical value of sensor data depends entirely on recovering resident-level context rather than merely detecting that some occupant triggered a sensor. This inability to isolate resident-specific event streams has effectively blocked the practical implementation of advanced services. Specifically, clinically meaningful monitoring of behavioral changes~\cite{aramendi2018automatic, giannios2024semantic, riboni2016smartfaber}, resident-specific anomaly detection~\cite{fahrmann2024anomaly, rahman2024comprehensive, rahim2023enhancing, afroz2025intrusion}, and personalized building automation~\cite{ahmadi2018real, lashkari2019energy, mahmud2024towards, wu2024smart} all remain largely unusable in shared homes because they cannot reliably attribute activities or preferences to a specific target individual. Without a robust mechanism to separate interleaved sensor streams and reconstruct individual trajectories, ambient intelligence remains confined to unrealistic single-occupant assumptions that fail to meet the personalization and reliability requirements of true residential deployment. More broadly, this challenge sits at the systems boundary between sensing and personalized interaction: ubiquitous computing applications can be embedded, privacy-preserving, and continuously available in the home, but they remain fundamentally limited if they cannot maintain resident-specific context under realistic multi-occupant conditions.

This technical gap stems from the inherent nature of ambient sensors, which typically provide sparse, fixed-location, and event-driven observations without explicit identity labels. When several residents share the same environment, their sensor events become temporally interleaved, thereby obscuring the continuity of each individual's behavior. The task of recovering separated individual trajectories from these interleaved observations is called Data Association~\cite{wilson2005simultaneous, wang2020smrt, arrotta2024multi, benmansour2015multioccupant, hsu2010strategies, chen2014two, chiang2010interaction, wang2021multi}. Compared with indoor localization, which focuses on continuously estimating coordinates~\cite{hayward2022survey}, Data Association operates on sensor events that are semantically richer but much sparser. This higher semantic level is particularly valuable for downstream applications because it is more directly tied to meaningful activities and interactions. For example, knowing that a resident is in the kitchen does not by itself reveal whether the resident is cooking or washing dishes. In contrast, observing that the stove's power sensor reports an increase provides stronger semantic evidence that the resident may be cooking. The goal of Data Association is therefore to disentangle these sparse symbolic events into coherent resident-level event streams so that downstream semantic inference, such as activity recognition~\cite{arrotta2024multi}, can be carried out on each recovered trajectory.

Existing Data Association methods largely fall into \textit{event-centric} and \textit{global state} paradigms, yet both face significant structural limitations. 
Event-centric approaches~\cite{crandall2010using, chen2024generative} classify observations with limited context. However, by lacking a persistent memory of ``silent'' residents, they fundamentally fail to maintain global consistency, often producing conflicting associations that violate multi-resident physical constraints. In practical shared-home settings, such inconsistencies can blur who performed which activity, directly undermining resident-specific interpretation.
Conversely, global state approaches attempt to track the full environment but struggle with modeling capability. 
Classic discrete filtering methods~\cite{wilson2005simultaneous} suffer from combinatorial explosion in the hypothesis space and are typically restricted to first-order Markovian transitions, limiting their ability to capture long-term behavioral dependencies. 
Continuous embedding approaches~\cite{wang2020smrt, wang2021multi} circumvent this combinatorial bottleneck by projecting discrete events into a continuous latent space to leverage efficient approximations (e.g., GM-PHD filters). 
However, this paradigm introduces an ontological mismatch: by enforcing linear dynamic model on discrete symbolic behaviors, they suffer from covariance divergence during sparse sensor activations, leading to trajectory fragmentation. In practice, these fragmented trajectories can break resident-level behavioral summaries and destabilize downstream monitoring or anomaly attribution.
Most recently, LLM-based approaches~\cite{chen2024towards} have revisited discrete reasoning but primarily rely on frozen pre-trained knowledge rather than actively utilizing sensor data for training, resulting in weak adaptation to domain-specific distributions. These limitations highlight the need for a principled framework that synergizes (i) an \textbf{extensible, history-augmented discrete state} that transcends first-order Markovian limits to capture long-term context, and (ii) \textbf{data-driven transition dynamics} capable of resolving complex association ambiguities via learned domain knowledge.

In this paper, we address these requirements by reformulating multi-resident Data Association as a \textbf{Markov Decision Process (MDP)}~\cite{puterman2014markov}. Our key shift is conceptual: instead of treating each sensor event as an isolated classification target or estimating assignments through recursive filtering alone, we cast the problem as sequential decision-making over a history-augmented global state. In our formulation, the state explicitly summarizes the evolving assignment history of all residents, and each action selects the resident set responsible for the current event. This perspective is well suited to shared homes because it enables the policy to incorporate long-range contextual dependencies, maintain logical consistency across residents, and naturally handle the dynamic evolution of household occupancy. Moreover, by explicitly organizing the task around states, actions, and sequential decisions, the MDP provides a principled framework for learning-based data association. In practice, however, specifying a reliable reward function for online deployment is difficult, whereas expert trajectories are already available in annotated datasets. We therefore solve the proposed MDP through \textbf{Behavior Cloning (BC)}~\cite{lin2024behavioral}, treating labeled event streams as demonstrations of expert assignment decisions. This yields a practical data-driven learning strategy while preserving the sequential semantics of the formulation. To study how different policy classes exploit this framework, we instantiate the policy in two complementary ways: a lightweight Transformer~\cite{vaswani2017attention} that learns temporal assignment patterns from scratch, and a fine-tuned Large Language Model (LLM)~\cite{zhao2023survey} that casts resident assignment as conditional generation and can better transfer across heterogeneous sensor environments.

We evaluate our proposed framework on three diverse public datasets: CASAS~\cite{singla2010recognizing}, MARBLE~\cite{arrotta2021marble}, and MuRAL~\cite{chen2025mural}, covering a wide range of multi-resident scenarios and sensing configurations. 
Extensive experiments show that our MDP-based approach, trained via Behavior Cloning, consistently outperforms state-of-the-art baselines~\cite{wang2020smrt, chen2024towards} by a substantial margin. In particular, it better maintains global consistency, assigns ambiguous events through beam search over candidate trajectories, and resolves concurrent multi-resident events that prior methods struggle to handle. Furthermore, the LLM-instantiated policy exhibits strong cross-environment generalization, demonstrating that semantic knowledge can transfer across heterogeneous sensor environments without extensive retraining. Finally, a simulated downstream health-monitoring analysis shows that these gains in data association also translate into substantially lower activity-duration estimation errors, further supporting the practical value of our method for resident-level health applications.
In summary, our main contributions are as follows:

\begin{itemize}
    \item We reformulate multi-resident Data Association as a history-aware MDP, providing a unified decision-making framework that explicitly models assignment continuity over time. This formulation moves beyond isolated event classification and first-order filtering, and makes it possible to address ambiguous and overlapping multi-resident events in a principled way.

    \item We develop a practical Behavior Cloning solution to this MDP and instantiate it with two complementary policy classes: a lightweight Transformer for efficient data-driven assignment and a fine-tuned LLM for stronger semantic transfer across environments. This design avoids reward engineering while supporting both accurate trajectory recovery and improved generalization under heterogeneous sensor settings.

    \item We validate the framework on three public datasets (CASAS, MARBLE, and MuRAL), where it consistently surpasses strong baselines across multiple metrics. Beyond benchmark gains, our results show tangible practical value: the method better preserves resident-level consistency, improves the handling of ambiguous and concurrent events, and reduces downstream activity-duration estimation errors in a simulated health-monitoring setting.
\end{itemize}

\section{Related Work}
\label{sect:relatedwork}
\subsection{Practical Limitations of Ambient-based Smart Home Applications}

Ambient sensors are an important modality for understanding residents' daily activities in smart homes. Their practical appeal stems from several well-known advantages: they are non-intrusive, low-power, privacy-preserving, and easy to deploy throughout the home. These characteristics also contribute to their user acceptability. Chan et al.~\cite{chan2024home} reported that in-home monitoring technologies show moderately positive acceptability overall, particularly for ambient sensors, and highlighted evidence that ambient sensors are better accepted than wearable alternatives. Precisely because of these advantages, ambient sensors have been widely adopted to support downstream applications such as health monitoring~\cite{aramendi2018automatic, giannios2024semantic, riboni2016smartfaber, chan2024home}, home safety and security~\cite{fahrmann2024anomaly, rahman2024comprehensive, rahim2023enhancing, afroz2025intrusion}, and building automation and energy management~\cite{ahmadi2018real, lashkari2019energy, mahmud2024towards, wu2024smart}. At the core of these applications lies the extraction of contextual information, with Activities of Daily Living (ADLs) serving as a particularly important semantic layer.

This dependence on ADL-level context is especially evident in healthcare-oriented smart-home systems, where the goal is continuous, resident-specific monitoring of clinically meaningful behavioral changes. Aramendi et al.~\cite{aramendi2018automatic}, for example, transformed ambient sensor streams into activity labels and derived indicators such as cooking duration, meal times, and nighttime bathroom frequency to predict functional health decline in older adults. Giannios et al.~\cite{giannios2024semantic} used ambient events and ADL information to quantify deviations in meal-preparation behavior for distinguishing healthy residents from those with mild cognitive impairment, while Riboni et al.~\cite{riboni2016smartfaber} combined activity classification and anomaly detection to support the early identification of cognitive decline. Yet all of these pipelines depend on \textit{individual-level} attribution: the system must know whose meals are becoming irregular, whose nocturnal bathroom visits are increasing, and whose routines are deviating from baseline. This dependence becomes a major obstacle in shared homes. Aramendi et al. explicitly discarded all two-resident samples because activities from different occupants could not be reliably separated; Giannios et al. similarly noted that the lack of individual-level ADL recognition confines their framework to single-resident settings; and Riboni et al. identified the separation of individual activities in multi-resident environments as necessary future work. At a broader level, Chan et al.~\cite{chan2024home} surveyed 46 studies on smart-home health monitoring and found that evidence for multi-person ambient sensing remains extremely limited: only a small minority reported any multi-person detection accuracy, the few ambient-sensor studies addressing multi-person settings focused mainly on basic presence or identity discrimination rather than clinically meaningful ADL-based outcomes, and no clinical study using ambient sensors provided dedicated validation in realistic shared-home settings. Collectively, these findings suggest that the main bottleneck is not the lack of clinical value, but the unresolved difficulty of reliably isolating each resident's activities in multi-resident homes.

A similar set of challenges plagues domestic safety and building automation. Recent surveys~\cite{fahrmann2024anomaly, rahman2024comprehensive} explicitly point out that existing anomaly-detection methods remain overwhelmingly anchored to single-occupant assumptions. As these studies emphasize, the inability to reliably separate individual event trajectories makes it difficult to analyze resident-specific anomalies (e.g., falls), distinguish genuine behavioral changes from multi-user interference, or detect external intrusions with confidence.

This limitation also extends to energy-saving applications~\cite{ahmadi2018real, mahmud2024towards, wu2024smart}, where ambient sensing is used to control lighting and heating, ventilation, and air conditioning (HVAC) systems. Lashkari et al.~\cite{lashkari2019energy} noted that most literature neglects realistic shared-home conditions involving concurrent behaviors and heterogeneous comfort preferences. Without clean trajectory separation, the system cannot personalize comfort or optimize energy load effectively: apparent vacancies may merely reflect hand-offs between occupants, and energy-control decisions may become inefficient or inappropriate. Consequently, across both safety and energy domains, the recovery of resident-level trajectories remains a fundamental barrier to practical deployment.

The collective evidence from these diverse applications points to a common practical limitation, namely the inability to distinguish between occupants in multi-resident settings. This constraint directly hinders the translation of promising research findings into widespread real-world deployment. Consequently, existing studies emphasize a pressing need for solutions capable of reliably attributing sensor events to specific individuals. This task is formally defined as Data Association~\cite{wilson2005simultaneous, wang2020smrt, arrotta2024multi, benmansour2015multioccupant, hsu2010strategies, chen2014two}. Rather than a mere technical refinement, Data Association represents the fundamental enabling step that allows healthcare, safety, and automation services to move beyond unrealistic single-occupancy assumptions. Prior empirical studies~\cite{chiang2010interaction, wang2021multi, chen2024generative} reinforce this by showing that robust data association can substantially improve multi-resident ADL recognition performance, thereby serving as a necessary foundation for deploying intelligent ambient services in realistic shared homes.

\subsection{Multi-Resident Data Association}\label{sec: sota_data_association}

Following a detailed review of Data Association methods developed over the past two decades, a clear methodological dichotomy emerges based on the scope of system state modeling: \textit{event-centric approaches}, which focus on characterizing individual observations in isolation or local context, and \textit{global state approaches}, which attempt to maintain a holistic representation of all residents in the environment.

\subsubsection{Event-Centric and Partial State Approaches}

Earlier approaches primarily treated multi-resident data association as a direct classification problem, minimizing the reliance on state maintenance. A classic example is the work by Crandall and Cook~\cite{crandall2008attributing, crandall2008resident}, who applied a Naive Bayes (NB) classifier to assign each sensor event to the most likely resident. However, this stateless approach is inherently limited by its inability to capture context. For instance, if the system assigns an event in the bedroom to Alice, a subsequent event in the living room is physically less likely to be Alice's. A stateless classifier, lacking memory of Alice's recent location, fails to utilize this spatial constraint. To address this, subsequent works introduced temporal modeling, yet they stopped short of a full system representation, adopting instead a \textit{partial state} formulation. Methods by Crandall et al.~\cite{crandall2010using} and Singla et al.~\cite{singla2010recognizing} utilized Hidden Markov Models (HMMs) where the hidden state represented only the identity or activity associated with the \textit{current} observation. Similarly, joint models~\cite{hsu2010strategies, cook2010detection} using CRFs~\cite{hsu2010strategies} or HMM~\cite{cook2010detection} inferred identity and activity simultaneously but remained fundamentally event-centric.

The critical limitation shared by these methods is that the state representation collapses to a single variable tuple associated only with the active resident triggering the sensor. Consequently, the system maintains no memory of residents who are currently silent (i.e., not triggering sensors). For example, if Alice is sleeping (silent) while Bob is cooking (active), these partial state models effectively lose track of Alice. By redefining ``state'' strictly as a hidden attribute of the current event rather than the configuration of the environment, these approaches functionally regress toward stateless classification, lacking the global view necessary to reason about temporal continuity for all residents simultaneously. This limitation extends even to modern deep sequence models like Seq2Res~\cite{chen2024generative}: while they expand the observation window from a single event to a sequence block, they still lack a persistent global memory, effectively treating the task as a batch translation problem that suffers from state forgetting over long durations.

\subsubsection{Global State Approaches}

Recognizing the necessity of tracking the full environment, Wilson and Atkeson~\cite{wilson2005simultaneous} established the foundational definition of \textit{Global State} modeling. They formalized the system state at time $t$ as the collection of all individual resident states:
\[
S_t = \{s^1_t, s^2_t, \dots, s^K_t\},
\]
where $s^i_t$ represents the discrete location and activity of resident $i$. 

To estimate this complex joint state, they proposed a Rao-Blackwellized Particle Filter (RBPF) framework. Crucially, this approach conceptually decomposes the multi-resident Data Association task into three sub-tasks: 
(1) \textbf{\textit{global state representation}}, which maintains the joint configuration of all residents; 
(2) \textbf{\textit{individual state transition}}, where each resident's dynamics are modeled as an independent HMM, allowing motion patterns to be learned from simpler single-resident data; and 
(3) \textbf{\textit{joint state hypothesis filtering}}, which uses particles to sample data associations and weight them against observations. 
This formulation enables a modular and semantically accurate approach to reasoning. However, despite its theoretical elegance, this approach suffers from a severe computational bottleneck: combinatorial explosion within the hypothesis filtering step. As the number of residents and sensor events increases, the discrete hypothesis space for data association grows exponentially. Consequently, an impractical number of particles is required to cover the valid association possibilities, making the approach computationally intractable for larger-scale deployments.

To overcome this barrier, recent methods such as sMRT~\cite{wang2020smrt} and GAMUT~\cite{wang2021multi} proposed a paradigm shift. Instead of modeling the system in its native discrete space, they project discrete sensor events into a continuous latent space via representation learning (e.g., Word2Vec). This transformation allows the global state to be defined as a set of continuous coordinate vectors $X_t = \{z^1_t, \dots, z^{K_t}_t\}$, enabling the use of the Gaussian Mixture Probability Hypothesis Density (GM-PHD) filter. The GM-PHD filter scales linearly with the number of targets and naturally handles time-varying resident numbers, effectively solving the computational tractability issue. Nevertheless, this shift to continuous embedding introduces a fundamental ontological mismatch. Human activity is inherently logical and symbolic, which is driven by intent (e.g., \textit{Bedroom} $\rightarrow$ \textit{Kitchen}) rather than by physical inertia. By enforcing a continuous representation, these methods impose artificial kinematic constraints (e.g., constant velocity) on discrete logical jumps. This theoretical flaw manifests mathematically as covariance divergence. Due to the temporal sparsity of smart home events, the error covariance matrix of the linear dynamic model grows monotonically during long intervals between observations. When the uncertainty exceeds the filter's threshold, valid trajectories are terminated, and subsequent events are mistakenly initialized as new tracks. Thus, while continuous methods solve the scalability problem, they do so at the cost of semantic fidelity, leading to \textbf{trajectory fragmentation} caused by applying physical motion models to discrete symbolic behaviors. In practical terms, this fragmentation is especially problematic for the applications discussed above: once a resident's trajectory is repeatedly broken or reassigned, downstream systems lose the continuity needed to interpret prolonged routines, attribute anomalies to the correct individual, or trigger personalized services with confidence in multi-resident homes.

Most recently, LAHAR~\cite{chen2024towards} has revisited the discrete global state paradigm, serving as a modern semantic successor to Wilson and Atkeson's framework. Instead of relying on computationally expensive particle filters, LAHAR leverages LLMs to explicitly represent the global state in natural language. By summarizing each resident's past state and using the LLM's reasoning capabilities to infer the next state $S_{t+1}$ given a new event, LAHAR effectively bypasses the combinatorial complexity of traditional filtering while preserving the logical fidelity of human behavior. However, while this approach resolves the semantic issues of continuous models, it introduces new challenges intrinsic to LLMs: prohibitive computational costs, high inference latency, and the under-utilization of available sensor training data, as the model relies more on pre-trained knowledge than on learning from domain-specific distributions.

\subsubsection{Synthesis: Towards a Decision-Making Paradigm}

Synthesizing these historical developments, our approach is fundamentally premised on a strict adherence to the discrete Global State formulation~\cite{wilson2005simultaneous}, which we regard as the prerequisite for maintaining multi-resident logical consistency. 
However, while we adopt this representation, we identify a critical structural limitation in the frameworks historically used to estimate it. 
Whether utilizing classical discrete methods (e.g., RBPF~\cite{wilson2005simultaneous}) or modern continuous embeddings (e.g., sMRT~\cite{wang2020smrt}), existing approaches predominantly operate within the Bayesian Filtering paradigm. 
Consequently, they are bound by the rigid First-Order Markov assumption, relying recursively on the immediate previous state to estimate the current one. 
This restriction inherently limits their ability to capture complex, long-term behavioral dependencies. 
Furthermore, each faces specific hurdles: discrete filters suffer from combinatorial explosion, while continuous filters incur an ontological mismatch by imposing physical motion models on logical behaviors.

To transcend these dual limitations—preserving the rigor of the global state while breaking the first-order barrier—we propose shifting the methodological perspective from recursive filtering to sequential decision-making. 
This insight naturally leads to our Markov Decision Process (MDP)~\cite{puterman2014markov} formulation. 
By explicitly augmenting the state with the assignment history, our framework enables the policy to reason about rich contexts, thereby combining the semantic fidelity of the global state with the computational power of data-driven learning.

\label{sect:methodology}
\section{Problem Definition and Method Overview}
\subsection{Problem Definition}
Prior work often uses the term \textit{Data Association} informally without a precise, task-level definition, which limits rigorous discussion and reasoning about objectives, assumptions, and evaluation.
To better illustrate our objectives and methodology, we therefore first provide a formal definition of the Data Association problem.

We first define a \textbf{\textit{sensory environment}} as a set $\mathrm{E} = \{o_i\}_{1\leq i \leq I}$, where each $o_i$ represents an ambient sensor along with its associated information (e.g., type, location, and capabilities) and $I$ is the total number of sensors in this environment.

Within a sensory environment, a set of \textbf{\textit{subjects}} $\mathrm{U} = \{u_k\}_{1 \leq k \leq K}$ can enter, exit, and perform various activities. Here, $K$ denotes the total number of distinct subjects that have appeared up to time $t$. As time evolves, $K$ is a non-decreasing function of $t$, since new subjects may appear but previously observed subjects remain counted.

A sensor event $e_t$ occurred in the environment $\mathrm{E}$ is defined as a tuple:
\begin{equation}
    e_t = (t, o_i, c),
\end{equation}

where:
\begin{itemize}
    \item $t$ is the timestamp at which the sensor state change occurs.
    \item $o_i \in \mathrm{E}$ is an ambient sensor in the environment.
    \item $c$ is the change of the sensor state (e.g., activation, deactivation, or a value update).
\end{itemize}

Given a sequence of sensor events recorded from the environment, denoted as $[e_{t}]_{1 \leq t \leq T}$, where $T$ is the length of this sequence, the \textbf{Data Association} problem consists in constructing a mapping $M$ which maps each event $e_t$ to its corresponding residents $\mathcal{U}_t$. Formally, we have:
\begin{equation}
    M_U: e_t \mapsto \mathcal{U}_t,
\end{equation}
where $\mathcal{U}_t \subseteq \mathrm{U}$ is a subset of the total subject set $\mathrm{U} = \{u_k\}_{1 \leq k \leq K}$.

\begin{figure}[t!]
    \centering
    \includegraphics[width=0.9\linewidth]{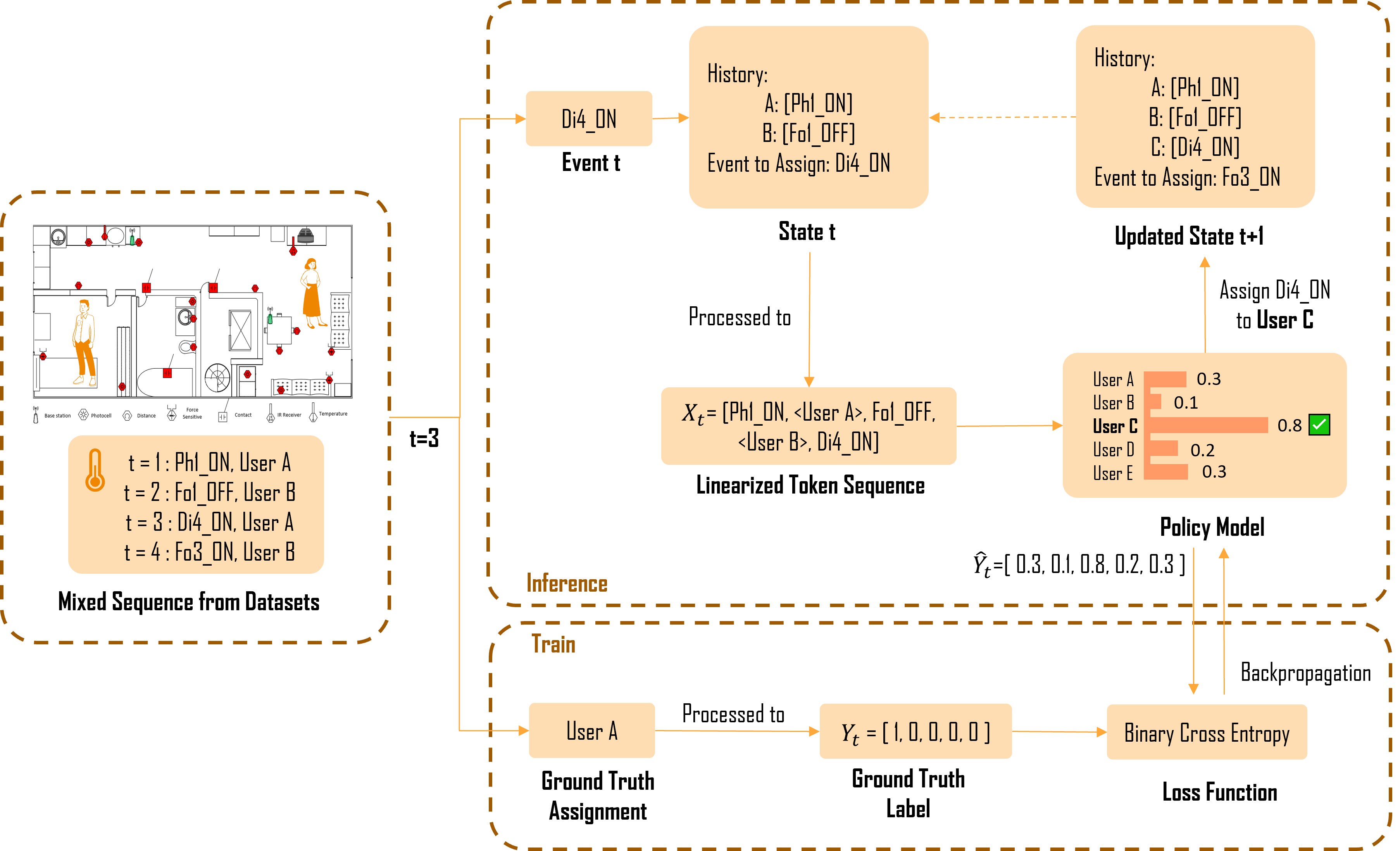}
\caption{Overview of our proposed MDP-based Data Association method, trained using Behavior Cloning.}
\label{fig:overview}
\end{figure}

Depending on the sensor modality and the nature of the event, $\mathcal{U}_t$ can be an empty set, a single subject, or multiple subjects. The empty-set case corresponds to events not attributable to any resident, which we uniformly treat as \textbf{noise}. Such noise may arise from sensor malfunction, measurement errors, or natural environmental changes (e.g., daily variations in temperature or illumination that trigger thermometer or light sensors). A single-subject case corresponds to actions such as one person opening a door, whereas multi-subject cases occur when several people interact within a shared space.

\textbf{Example:} To illustrate the above definitions, we consider a simple scenario: Alice opens the bedroom door and walks out, while Bob stands up from a chair. Then, both Alice and Bob walk together to the apartment's main door and open it. Suppose there are three sensors installed in the environment: a bedroom door sensor \( D_1 \), a chair pressure sensor \( C_1 \), and a main door sensor \( D_2 \). The environment can thus be defined as: $\mathrm{E} = \{ D_1, C_1, D_2 \}$. In this scenario, three sensor events occur in sequence: $[e_1 = (1, D_1, \text{OPEN}), e_2 = (2, C_1, \text{OFF}), e_3 = (3, D_2, \text{OPEN})]$. The goal of Data Association is to construct a mapping \( M \) from events to residents such that: $M(e_1) = \{Alice\}, M(e_2) = \{Bob\}, M(e_3) = \{Alice, Bob\}$.

\subsection{Method Overview}

Figure~\ref{fig:overview} illustrates the detailed workflow of our proposed framework. 
The process begins with the MDP Formulation (Section~\ref{sec: mdp}), where the system constructs a structured state containing the resident history and the current event. 
To process this symbolic state, we process it into a linearized token sequence $X_t$. 
The policy model is instantiated as a Transformer network (Section~\ref{sec: network_arch}) to generate assignment probabilities $\widehat{Y}_t$. 
In the Inference loop, the selected action updates the global history, transitioning the system to the next state. 
In the Training loop, the framework follows the Behavior Cloning paradigm (Section~\ref{sec: bc}): it utilizes ground truth assignments $Y_t$ to compute the Binary Cross Entropy loss, optimizing the policy model via backpropagation. 
In addition to the architecture depicted here in Figure~\ref{fig:overview}, we further implement an alternative policy based on LLMs, which reformulates the task as conditional text generation (Section~\ref{sec:llm_policy}).

\section{Problem Modeling via MDP}\label{sec: mdp}
In this section, we present how the Data Association problem can be formally modeled as an MDP. We first motivate this formulation in Section \ref{sec: mdp_rationale} by discussing the decision-making and sequential properties of the task. In Section \ref{sec: mdp_def}, we then provide the basic definition of an MDP. Finally, in Section \ref{sec: adapt_mdp_to_da}, we formally describe how to model Data Association as an MDP, laying the foundation for solving the problem through MDP-based approaches in the subsequent sections.

\subsection{Rationale for the MDP Formulation}\label{sec: mdp_rationale}

We adopt the MDP formulation because the Data Association problem exhibits the following properties:

\begin{enumerate}
    \item \textbf{Data Association is fundamentally a decision-making problem.} Data Association is not simply a real-time filtering problem. Filtering usually focuses only on estimating the current state, whereas Data Association often involves long-horizon decision-making. For example, when two residents pass through the main door at the same time, an anonymous \texttt{Main Door OPEN} event alone may be insufficient to determine whether the event should be assigned to one resident or to both. The system may need to postpone commitment in the hypothesis space and reason about future possibilities, such as subsequent sensor activations in different rooms, before it can retrospectively determine the current assignment. This property of using future evidence to inform the present gives Data Association a fundamentally decision-making character.

    \item \textbf{Data Association is inherently sequential.} Assignment decisions exhibit a strong chain effect. Because of physical continuity and the logical structure of daily routines, the association made at the current step directly constrains the space of plausible assignments at subsequent steps. For example, if the system determines that a \texttt{Bathroom Door CLOSE} event is triggered by a given resident, then under spatiotemporal consistency the immediately following \texttt{Bathroom Door OPEN} event should be much more likely to be assigned to the same resident. This direct shaping of future decisions by earlier actions is a fundamental sequential property of the problem.

    \item \textbf{These properties naturally align with the MDP framework.} Taken together, the sequential decision-making nature of Data Association matches the mathematical structure of a Markov Decision Process. Its decision-making aspect corresponds to the ability of an MDP to search for an optimal policy that maximizes long-term return, while its sequential dependence corresponds to the way state transitions dynamically constrain future actions. By modeling Data Association as an MDP, we obtain a principled formalism for representing resident history, current observations, and the long-term consequences of present assignments in a unified framework.
\end{enumerate}

\subsection{Definition of MDP} \label{sec: mdp_def}
An agent is formally defined as an entity that makes decisions and takes actions within an environment, aiming to maximize long-term cumulative reward. An MDP is a mathematical framework that models the interaction between an agent and its environment in sequential decision-making settings.

A typical MDP can be defined by a tuple \((S, A, P, R)\), where 
\begin{itemize}
    \item \(S\) is the state space, representing all possible states of the agent,
    \item \(A\) is the action space, representing all actions that the agent is allowed to execute,
    \item \(P(s'|s, a)\) is the state transition model, the probability of reaching state $s'$ from state $s$ by taking action $a$,
    \item \(R(s, a, s')\) is the reward model, the immediate reward for transitioning from $s$ to $s'$ via action $a$.
\end{itemize}
The agent's objective is to optimize a decision policy $\pi: S \to A$, which maps states to actions, such that the expected cumulative reward over a sequence of actions is maximized:
\begin{equation}\label{eq: pi_star}
    \pi^* = \arg\max_{\pi} \mathbb{E} \left[ \sum_{t=0}^{T} R(s_t, \pi(s_t), s_{t+1}) \right],
\end{equation}
where $t$ is the index of discrete time step and $T$ is the maximum of time steps. In this study, we only consider a finite number of decision steps.

\subsection{Adapting MDP to Data Association} \label{sec: adapt_mdp_to_da}
Under the MDP framework, the Data Association problem is modeled by the following components:

\paragraph{\textbf{State}}
At any event step $t$, the state $s_t$ encodes the assignment history prior to \( t \), denoted as \( h_{<t} \), along with the current event \( e_t \) to be assigned. Formally, we define the state as:
\begin{equation}
    s_t = (h_{<t}, e_t).
\end{equation}
Here, \( h_{<t} \) serves as a dynamic record, implemented as a mapping from each resident to their assigned event sequence (cf. state example in Figure~\ref{fig:overview}). This state representation captures both past context and the current decision point, and thus satisfies the Markov property.

\paragraph{\textbf{Action}}

As shown in the example of Figure~\ref{fig:overview}, at each step \( t \), the agent takes an action \( a_t \), assigning the current event \( e_t \) to a non-empty subset of residents:
\begin{equation}
    a_t: e_t \mapsto \widehat{\mathcal{U}}_t \subseteq \mathbb{U}_t, \quad \widehat{\mathcal{U}}_t \neq \emptyset,
\end{equation}
where \( \mathbb{U}_t = \{ u_1, \dots, u_n, u_{\text{new}} \} \) includes all known residents and a placeholder \( u_{\text{new}} \) for dynamically introducing new ones.

This formulation supports both single- and multi-resident assignments, enabling the model to handle overlapping activities and adapt to scenarios with an unknown or growing number of residents.

\paragraph{\textbf{Transition Model}}

Given a state \( s_t = (h_{<t}, e_t) \) and an action \( a_t \), the environment deterministically transitions to the next state by updating the assignment history and moving to the next event. Specifically, the current event \( e_t \) is appended to the event list of each assigned resident in \( \widehat{\mathcal{U}}_t = a_t(e_t) \), yielding a new history \( h_{<t+1} \) (cf. the updated state in Figure~\ref{fig:overview}). If \( \widehat{\mathcal{U}}_t \) includes a resident not previously seen in \( h_{<t} \), a new resident is created accordingly and labeled using the next available alphabetical ID.
 The next unprocessed event \( e_{t+1} \) becomes the new focus of assignment. Formally:
\begin{equation}
    s_{t+1} = (h_{<t+1}, e_{t+1}).
\end{equation}

This transition is deterministic and satisfies the Markov property: all necessary information for future decisions is encapsulated in the updated state.

\paragraph{\textbf{Reward}}

The reward function quantifies the immediate correctness of the agent's decision at each step. To strictly align the MDP framework with the objective of accurate tracking, we define an oracle reward using the indicator function:
\begin{equation}
    R(s_t, a_t) = \mathds{1}(a_t = a^*_t),\label{eq: oracle_reward}
\end{equation}
where $ a^*_t $ denotes the ground-truth resident assignment for event $ e_t $. Under this binary formulation, the agent receives a reward of $1$ solely when the assigned resident set matches the ground truth exactly, and $0$ otherwise. This definition establishes the fundamental metric for evaluating individual transitions.

\paragraph{\textbf{Policy}}

A \textit{policy} defines a strategy for Data Association, mapping the current state \( s_t \) to an assignment action \( a_t \). Formally, we denote this deterministic mapping as \( \pi(s_t) = a_t \). Executing a policy \( \pi \) over a sequence of sensor events \( [e_t]_{1 \leq t \leq T} \) induces a complete assignment trajectory.

The ultimate goal of Data Association is to recover the ground-truth mapping \(M\), which corresponds to an optimal policy \( \pi^* \) that assigns the correct resident set at every step:
\begin{equation}
    \forall t \in \{1, \dots, T\}, \quad M(e_t) = \pi^*(s_t) = a^*_t.
\end{equation}

\textbf{Consequently, the Data Association problem can be cast as a policy optimization task. }The objective is to find a policy \( \pi^{\text{opt}} \) that approximates \( \pi^* \) by maximizing the expected cumulative reward:
\begin{equation}
    \pi^{\text{opt}} = \arg\max_{\pi} \mathbb{E} \left[ \sum_{t=1}^{T} R(s_t, \pi(s_t)) \right].
\end{equation}

Since the reward $R$ is defined as the binary correctness of each step, maximizing the cumulative reward is mathematically equivalent to maximizing the total assignment accuracy over the event stream. This formulation provides a rigorous theoretical basis for seeking the optimal association strategy.

\section{Policy Learning via Behavior Cloning} \label{sec: bc}

\subsection{Rationale for Behavior Cloning}
Having formulated Data Association as an MDP in the previous section, the goal of this section is to solve the resulting decision problem in practice. In principle, many solution strategies are possible, including explicit search methods such as Monte Carlo Tree Search (MCTS), standard Reinforcement Learning (RL), and imitation-based policy learning. As the first work, to our knowledge, to cast multi-resident Data Association explicitly into this MDP form, our immediate objective is not to exhaust this algorithmic space, but to validate the formulation itself and establish a strong, transparent baseline for future work. We therefore seek a solver that is reliable, easy to interpret, and suitable for revealing both the strengths and the limitations of the proposed state--action--transition interface.

To this end, we adopt BC as the optimization strategy for the proposed MDP. Crucially, the assignment labels in annotated smart-home datasets already specify, at each decision step, the action that a human expert would regard as the correct attribution under full supervision. In other words, the dataset directly provides expert demonstrations of how the MDP should be solved. Rather than rediscovering this policy indirectly through online trial-and-error, we can therefore learn a parameterized policy $\pi_\theta$ by directly imitating these expert actions, which is the most data-efficient and conceptually direct strategy for our setting. This choice allows us to validate the MDP formulation in a controlled and interpretable manner, while avoiding confounding the contribution of the formulation with the additional instability of reward engineering and exploration. Furthermore, as is common practice across many domains of sequential decision-making, utilizing BC to provide a crucial ``warm start'' is often necessary before RL can efficiently explore vast combinatorial action spaces. Thus, our BC approach serves as an essential first step, providing a solid foundation and a strong baseline upon which more advanced RL- or search-based solvers can be investigated in future work.

\subsection{Learning Objective}

To implement the Behavior Cloning framework, we translate the MDP objective, which is maximizing the cumulative reward (i.e., assignment accuracy), into a supervised optimization task. Since the reward is binary, maximizing the expected reward is equivalent to maximizing the probability that the agent selects the ground-truth action $a^*_t$.

Therefore, we aim to optimize the policy parameters $\theta$ such that the probability mass assigned to the expert's actions is maximized. This corresponds to maximizing the log-likelihood of the observed expert trajectories under the learned policy distribution:
\begin{equation} \label{eq:bc-mle}
    \theta^* = \arg\max_{\theta} \sum_{t=1}^{T} \log P_\theta(a_t = \mathcal{U}_t \mid s_t).
\end{equation}

\paragraph{\textbf{Action Space Instantiation}}
While our general MDP formulation (Section~\ref{sec: adapt_mdp_to_da}) supports dynamic resident discovery via a growing set $\mathbb{U}_t$, implementing the policy using standard neural architectures requires a fixed output dimension. Therefore, for this learning instantiation, we define a fixed universal resident pool $\mathbb{U} = \{u_1, \dots, u_K\}$, where $K$ represents the maximum capacity of trackable residents in the environment.
Consequently, the action $a_t$ is restricted to selecting a subset from this pre-defined universe $\mathbb{U}$. This simplification creates a consistent label space for the multi-label classifier while remaining sufficient to cover all residents present in the target datasets.

\paragraph{\textbf{Optimization}}
To make the optimization tractable, we decompose the set-level prediction into individual resident probabilities. We assume that the assignment probability of each resident in the fixed pool $\mathbb{U}$ is conditionally independent given the state $s_t$. This assumption is justified because the state representation $s_t$ (comprising the event and assignment history) is designed to implicitly encode the complex inter-dependencies and context. Consequently, the log-likelihood factorizes as:
\begin{equation} \label{eq:bc-factorization}
    \log P_\theta(\mathcal{U}_t \mid s_t) = \sum_{u_k \in \mathcal{U}_t} \log P_\theta(u_k \mid s_t) + \sum_{u_k \notin \mathcal{U}_t} \log (1 - P_\theta(u_k \mid s_t)).
\end{equation}

We formalize this using binary labels $y_{t,k} = \mathds{1}(u_k \in \mathcal{U}_t)$, where $y_{t,k}=1$ indicates that resident $u_k$ is responsible for the event at step $t$, and $0$ otherwise. Substituting Eq.~\ref{eq:bc-factorization} into the objective yields the multi-label Binary Cross-Entropy (BCE) loss:
\begin{equation} \label{eq:bce}
    \mathcal{L}(\theta) = - \sum_{t=1}^{T} \sum_{k=1}^{K} \Big[ y_{t,k} \log P_\theta(u_k \mid s_t) + (1 - y_{t,k}) \log (1 - P_\theta(u_k \mid s_t)) \Big].
\end{equation}

Minimizing $\mathcal{L}(\theta)$ trains the network to act as a set of $K$ parallel binary classifiers. During inference, the simplest decoder constructs the predicted assignment set $\widehat{\mathcal{U}}_t$ by selecting all residents whose predicted probability $P_\theta(u_k \mid s_t)$ exceeds a decision threshold $\tau$ (e.g., $\tau=0.5$). This threshold-based mechanism naturally aligns with the MDP action space, enabling the agent to output flexible resident combinations based solely on learned confidence scores, ranging from empty sets representing noise to multi-resident groups. We later revisit this decoding rule in Section~\ref{sec: beam_search}, where we promote it from a one-shot per-step decision to a search over assignment trajectories.

\begin{figure}[t]
    \centering
    \includegraphics[width=0.85\textwidth]{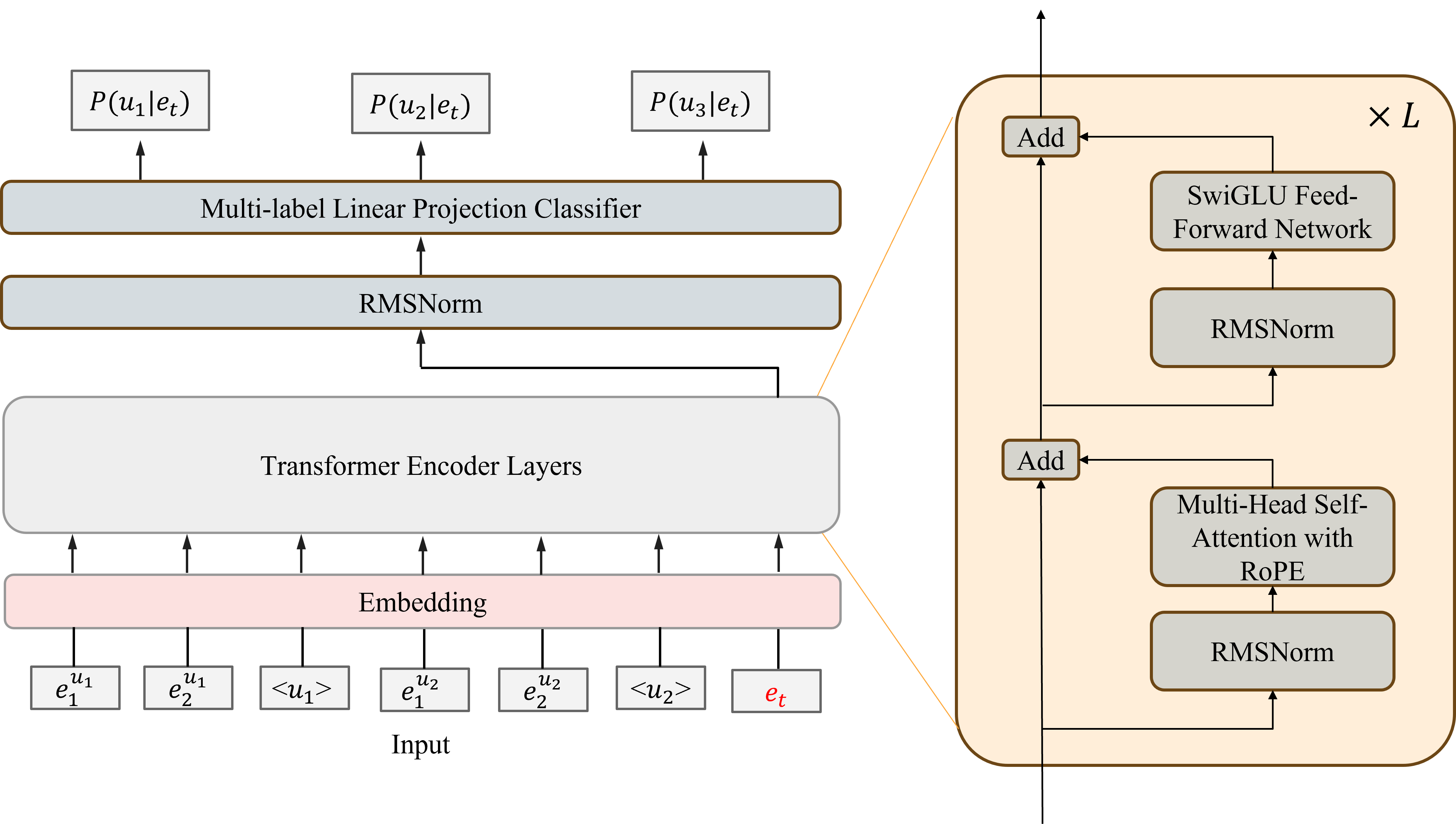}
    \caption{Architecture of the Transformer-based policy network. The model processes the linearized event sequence and outputs resident assignment probabilities via a multi-label classification head.}
    \label{fig:policy_model}
\end{figure}

\subsection{Policy Network Architecture} \label{sec: network_arch}

To implement the policy $\pi_\theta$, we design a neural network that maps the structured MDP state into resident assignment probabilities. As illustrated in Figure~\ref{fig:policy_model}, the architecture follows a three-stage processing pipeline: \textit{State Encoding}, \textit{Context Modeling}, and the \textit{Policy Head}.

\paragraph{\textbf{State Encoding}}
The first module converts the complex, multi-resident state $s_t = (h_{<t}, e_t)$ into a unified tensor representation. We employ a linearization strategy to flatten the hierarchical structure while preserving semantic boundaries.

Let $\mathbb{U}$ be the fixed universe of $K$ potential residents. At step $t$, with $n$ identified residents, we construct a token sequence $X_t$ by concatenating the local event history of each resident followed by the current event $e_t$. To distinguish between resident streams, we insert special delimiter tokens $\texttt{<}u_k\texttt{>}$. The resulting sequence is:
\[
    X_t = \big[
        e^{u_1}_1, \dots, e^{u_1}_L, \texttt{<}u_1\texttt{>}, \quad
        \dots, \quad
        e^{u_n}_1, \dots, e^{u_n}_L, \texttt{<}u_n\texttt{>}, \quad
        e_t
    \big].
\]
Here, $L$ is the context window size. This sequence is passed to an embedding layer, which projects discrete tokens (sensor events, delimiters, and special markers) into a continuous vector space. These dense embeddings serve as the input to the subsequent sequence model.

\paragraph{\textbf{Context Modeling}}
The core backbone is a stack of Transformer encoder layers designed to capture temporal and semantic dependencies. To maximize training stability and representation power, we adopt a modern Transformer variant incorporating Pre-RMSNorm, SwiGLU activations, and Rotary Positional Embeddings (RoPE). These architectural enhancements enable the model to effectively encode relative temporal distances and complex feature interactions. Through the self-attention mechanism, the current event token $e_t$ queries the entire history sequence, aggregating relevant behavioral cues from all identified residents to form a comprehensive decision context.

\paragraph{\textbf{Policy Head}}
The final module maps the contextualized features to the assignment action space. We extract the hidden state corresponding to the last token position (the current event $e_t$) as the summary representation $h_t$. This vector encapsulates the global context necessary for the decision.

We apply a final RMSNorm to $h_t$ and pass it through a linear projection layer to produce logits for all $K$ defined resident indices. The probabilities are computed via a sigmoid activation:
\[
    P_\theta(u_k \mid s_t) = \sigma(W_k^\top h_t + b_k), \quad k \in \{1, \dots, K\}.
\]
To handle the dynamic nature of resident discovery within this fixed topology, we interpret the outputs logically: indices $1$ to $n$ correspond to known residents, and index $n+1$ represents a potential new resident. Outputs at indices $k > n+1$ are masked during inference. This design allows the agent to output flexible resident combinations—including single assignments, multi-resident groups, or new resident discoveries—based strictly on the learned confidence scores.

\subsection{Training Data Curation} \label{sec: bc_data_aug}

\begin{figure}[t!]
    \centering
    \includegraphics[width=0.6\linewidth]{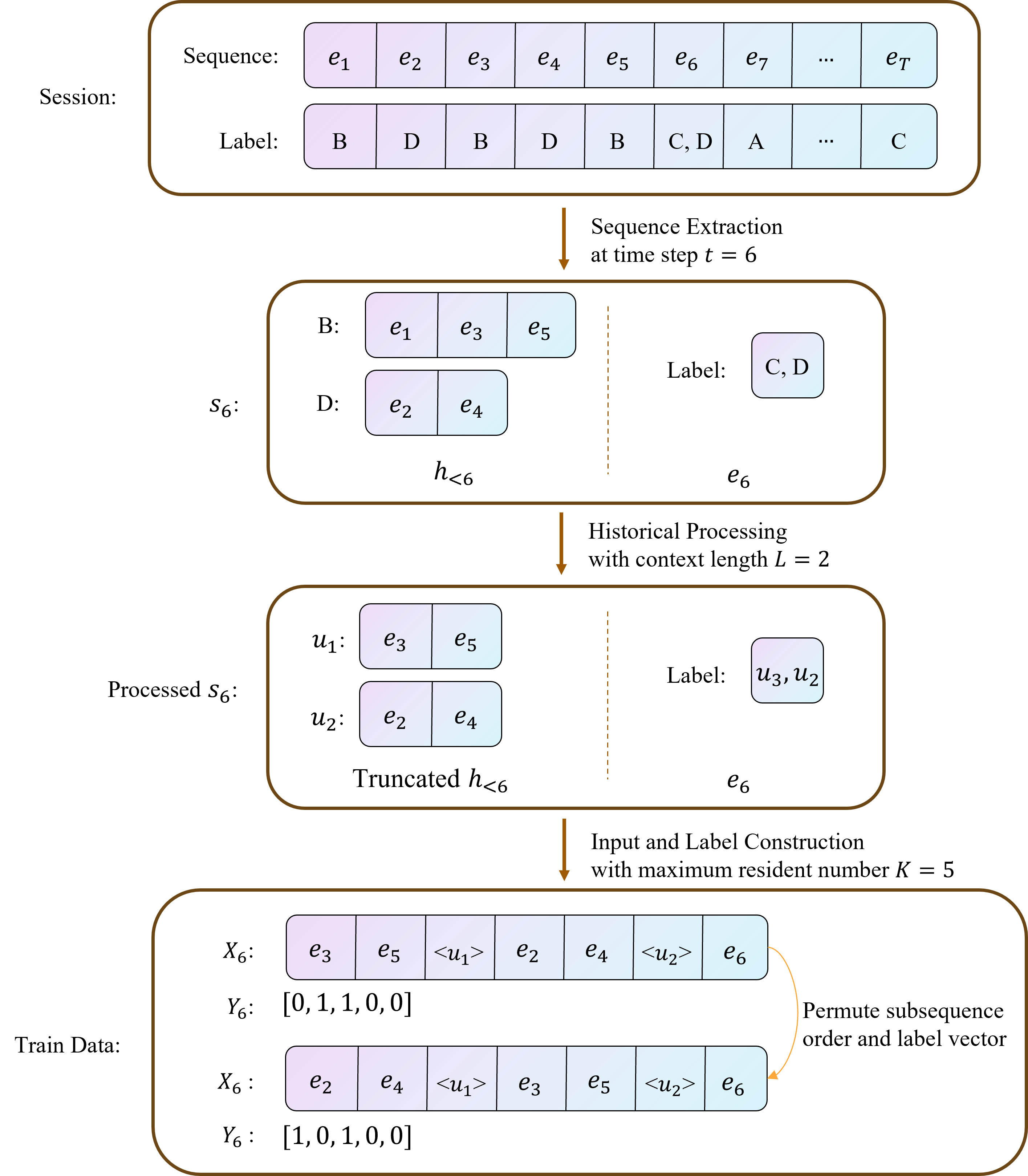}
    \caption{The training data preparation pipeline, illustrating sequence extraction, resident canonicalization, and permutation augmentation.}
    \label{fig:bc_train_data}
\end{figure}

We curate a supervised dataset by transforming raw sensor logs into structured history-label pairs. As illustrated in Figure~\ref{fig:bc_train_data}, this process involves sequence extraction, resident canonicalization, and permutation-based augmentation to ensure model robustness.

\paragraph{\textbf{Sequence Extraction}}
We process the chronological event stream using a sliding window approach. For a session $S = [e_1, \dots, e_T]$, we generate a training instance for every time step $t$. Events $e_{1:t-1}$ constitute the available history, while $e_t$ serves as the prediction target. This creates a dense set of examples reflecting the evolving context at every decision point.

\paragraph{\textbf{Resident Canonicalization and Grouping}}
To ensure the model learns generalizable behavioral patterns rather than memorizing specific resident IDs, we map residents to a canonical index space. For each instance at step $t$, we identify the subset of $n$ residents present in the history. These residents are mapped to local indices $\{1, \dots, n\}$ based on their order of appearance or an arbitrary sequence. We then group historical events by resident and truncate each subsequence to the most recent $L$ events.

\paragraph{\textbf{Permutation Augmentation}}
A critical challenge in using sequence models for multi-resident data is that the set of residents has no intrinsic order, yet the input $X_t$ requires a sequential concatenation. This introduces an artificial positional bias. To enforce \textit{permutation invariance}, we employ data augmentation.
For each training instance, we generate multiple variations by enumerating random permutations of the $n$ resident blocks in the input sequence. The target label vector is permuted accordingly. This strategy explicitly trains the model to associate the current event $e_t$ with specific behavioral histories regardless of their position in the input tensor, thereby achieving the desired permutation-equivariant behavior.

\paragraph{\textbf{Label Construction}}
The target label is encoded as a $K$-dimensional multi-hot binary vector. The $i$-th entry is set to 1 if the current event belongs to the resident mapped to canonical index $i$. If the event originates from a resident not present in the history (a new resident), we activate the $(n+1)$-th entry. This representation supports multi-label supervision and aligns with the dynamic resident discovery logic of the network head.

\paragraph{\textbf{Token Indexing}}
Finally, all symbolic tokens, including sensor events and resident delimiters, are mapped to integers using a global vocabulary. This vocabulary encompasses all distinct sensor states and delimiter types, producing the final tensor indices for the embedding layer.

\subsection{Inference via Beam Search over Assignment Trajectories} \label{sec: beam_search}

As argued in Section~\ref{sec: mdp}, Data Association differs from pure filtering because the current assignment may remain ambiguous and only become clear after several future events provide disambiguating context. Relying solely on the greedy decoder is still insufficient: by committing to a per-step argmax, it discards alternative hypotheses before future evidence arrives. To preserve this deferred-commitment capability, we reformulate inference as a search over assignment trajectories $\mathbf{a} = (a_1, \ldots, a_T)$ and use beam search as a tractable approximation that keeps the top-$B$ partial trajectories ranked by accumulated likelihood. When $B{=}1$, beam search reduces to the original greedy decoder; when $B{>}1$, it retains multiple plausible hypotheses that can be revised as later events arrive, without modifying the learned policy $\pi_\theta$.

Formally, at step $t$, the multi-label policy yields a per-slot Bernoulli probability $p_t^{(k)} = P_\theta(u_k \mid s_t)$. By the same conditional-independence factorization as in Eq.~\ref{eq:bc-factorization}, the joint probability of any candidate action $a_t \subseteq \mathbb{U}$ can be written as
\begin{equation} \label{eq:joint_action_prob}
    P_\theta(a_t \mid s_t) \;=\; \prod_{u_k \in a_t} p_t^{(k)} \;\cdot\!\!\!\prod_{u_k \notin a_t}\!\!\bigl(1 - p_t^{(k)}\bigr),
\end{equation}
and a complete trajectory is scored by its accumulated log-likelihood $S(\mathbf{a}) = \sum_{t=1}^{T} \log P_\theta(a_t \mid s_t)$. At each step, every surviving partial trajectory in the current beam expands its frontier by enumerating the non-empty subsets of slots whose marginal probability $p_t^{(k)}$ exceeds a pruning threshold $\tau$ (which keeps the branching factor bounded). Each child applies a deterministic state update that appends $e_t$ to the history of every assigned resident and accumulates the corresponding $\log P_\theta(a_t \mid s_t)$ into its score, and the global frontier is then truncated to the top-$B$ partial trajectories ranked by $S(\cdot)$. After all $T$ events have been processed, the trajectory with the highest accumulated score is returned as the final attribution.

\section{LLM as a Generative Policy} \label{sec:llm_policy}

While the scratch-trained architecture described in Section~\ref{sec: network_arch} effectively captures statistical dependencies, it is inherently limited by its reliance on dataset-specific sensor IDs. Such models tend to overfit to the fixed topology of the training environment and struggle to generalize to unseen interaction patterns or heterogeneous sensor layouts. To overcome this limitation, we propose an alternative policy implementation leveraging Large Language Models (LLMs). By grounding the data association task in natural language, we exploit the pre-trained world knowledge of LLMs to achieve robust generalization beyond mere statistical correlation. For instance, the model can leverage the semantic relationship between a motion sensor in the kitchen and cooking activities without requiring extensive training from scratch.

\begin{figure}[t!]
    \centering
    \includegraphics[width=0.9\linewidth]{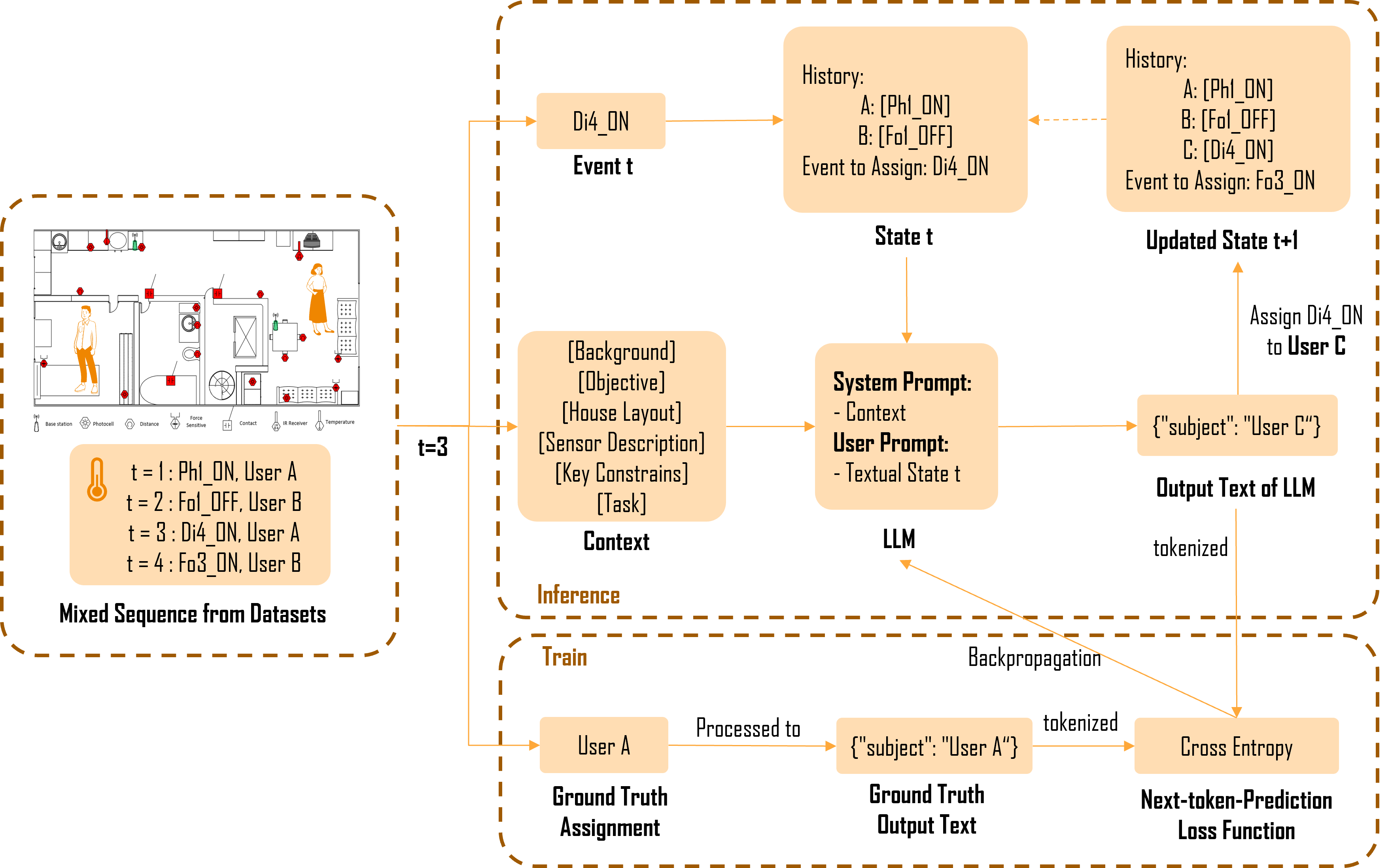}
    \caption{Overview of the LLM-based policy architecture. The discrete MDP state is serialized into a textual prompt, and the model is fine-tuned to generate the target resident assignment directly.}
    \label{fig:llm_bc}
\end{figure}

Figure~\ref{fig:llm_bc} illustrates our proposed framework. We reformulate the problem from a multi-label classification task into a supervised conditional text generation task. The workflow begins by serializing the environmental priors and interaction history into a semantic prompt. This context conditions the LLM to directly generate the resident assignment in a concise JSON format. Through this paradigm, the model learns to associate events based on universal semantic logic rather than rigid ID matching. This significantly enhances its ability to handle ambiguity and dynamic environmental changes.

\subsection{Generative Policy Learning}

To enable the model to comprehend the physical environment and generalize across heterogeneous sensor layouts, we augment the dynamic MDP state $s_t$ with static environmental context $c$. This context includes invariant priors such as the house layout and sensor descriptions. Adhering to the generative paradigm, we serialize both the state $s_t$ and context $c$ into a composite textual prompt $\mathcal{X}_t$. The policy $\pi_\theta$, parameterized by the LLM, learns to map this prompt to a target sequence $\mathcal{Y}_t = \{y_{t,1}, \dots, y_{t,L}\}$ that explicitly encodes the resident assignment decision in a structured JSON format (e.g., \texttt{\{"subject": "User A, User B"\}}).

Consistent with the Behavior Cloning framework established in Section~\ref{sec: bc}, we treat the Supervised Fine-Tuning (SFT) of the LLM as a direct policy imitation process. Adhering to the generative paradigm, we serialize both the state $s_t$ and context $c$ into a composite textual prompt $\mathcal{X}_t$. The policy $\pi_\theta$, parameterized by the LLM, learns to map this prompt to a target sequence $\mathcal{Y}_t = \{y_{t,1}, \dots, y_{t,L}\}$ that explicitly encodes the resident assignment decision in a structured JSON format (e.g., \texttt{\{"subject": "User A, User B"\}}).

In this framework, the policy defines a conditional distribution over the target token sequence:
\begin{equation}
    \pi_\theta(a_t \mid s_t) \equiv P_\theta(\mathcal{Y}_t \mid \mathcal{X}_t).
\end{equation}
To make optimization tractable, we decompose this joint probability autoregressively:
\begin{equation}
    P_\theta(\mathcal{Y}_t \mid \mathcal{X}_t) = \prod_{i=1}^{L} P_\theta(y_{t,i} \mid \mathcal{X}_t, y_{t,<i}),
\end{equation}
where $y_{t,<i}$ denotes the history of generated tokens for the current response.

We optimize the model parameters $\theta$ by minimizing the negative log-likelihood (NLL) of the generated tokens, corresponding to the standard Next Token Prediction (NTP) objective. This objective is mathematically equivalent to minimizing the divergence between the learner and expert distributions as defined in classical Behavior Cloning:
\begin{equation} \label{eq:ntp_loss}
    \mathcal{L}(\theta) = - \sum_{t=1}^{T} \sum_{i=1}^{L} \log P_\theta(y_{t,i} \mid \mathcal{X}_t, y_{t,<i}).
\end{equation}

During training, we employ teacher forcing by providing the ground-truth prefix $y_{t,<i}$ to predict the next token. Crucially, a loss mask is applied to the prompt tokens $\mathcal{X}_t$. This ensures that gradients are back-propagated exclusively from the prediction of the target resident assignment.

\subsection{Prompt Engineering}
To enable the language model to process sensor data, we employ a structured serialization strateg, mapping the numerical state $s_t$ and context $c$ into a structured textual input $\mathcal{X}_t$. Similar to general prompt designs, the schema consists of a static system prompt and a dynamic user prompt.

\paragraph{\textbf{User Prompt}}
The user prompt represents the dynamic state $s_t$. As shown in our prompt design, it serializes the current sensor reading $e_t$ (including timestamp, sensor ID, and location) and the recent history of residents $h_{<t}$ into a human-readable format. It further instructs the LLM to output the final assignment in a strict JSON format, so that our system can automatically parse the response and update the state. Taking the MARBLE dataset as an example, the final user prompt used in this study is shown as following:

  \begin{tcolorbox}[
  colback=myGreen!5,
  colframe=myGreen!75,
  title=User Prompt for MARBLE dataset used in our proposed method,
  fonttitle=\bfseries,
  breakable,
  enhanced
  ]

\small
Observed users:
\begin{itemize}[leftmargin=*, noitemsep, topsep=0pt]
    \item User\_1 (latest 1 events): \\
    {[12:14:44] P4 ON}
\end{itemize}

You can assign this event to:
\begin{enumerate}[leftmargin=*, noitemsep, topsep=0pt]
    \item An existing user (from the list above)
    \item A new user (User\_2) if this event belongs to someone not yet observed
\end{enumerate}

Current sensor event observation:
\begin{tcolorbox}[colback=white, colframe=gray!30, boxrule=0.2mm, arc=0mm, left=1mm, top=1mm, bottom=1mm]
\begin{verbatim}
{
  "time": "12:14:44",
  "event": "P1 ON",
  "sensor": "P1",
  "action": "ON",
  "location": "dining room"
}
\end{verbatim}
\end{tcolorbox}

Based on the historical patterns and location constraints, which user(s) triggered this event?

Output your prediction in JSON format: \{\{"subject": "User\_X"\}\} or \{\{"subject": "User\_X, User\_Y"\}\} for multiple users.

\end{tcolorbox}

\paragraph{\textbf{System Prompt}}
The system prompt encodes the static context $c$,, which provides static background knowledge that remains fixed across all inference steps. It defines the environment in which reasoning takes place, including:  
(1) the general background and objective of the task;  
(2) the spatial layout of the smart home and inter-room connectivity;  
(3) the mapping between sensor identifiers and physical objects; and  
(4) key behavioral and physical constraints that govern user mobility and spatial exclusivity.    
This structured global context serves as a static prior that grounds the model’s reasoning in the spatial, semantic, and physical realities of the environment. Taking the MARBLE dataset as an example, the final system prompt use in this study is shown as following on next page.

\clearpage
\begin{tcolorbox}[
  colback=myGreen!5,
  colframe=myGreen!75,
  title=System Prompt for MARBLE dataset used in our proposed method,
  fonttitle=\bfseries,
  breakable,
  enhanced
  ]

\small
\textbf{[Background]} \\
You are in a smart home where some sensors are installed. There are at most four people living in the house. They engage in daily activities at home. Their actions in the house can trigger the sensors and be recorded as sensor events. Patterns in sensor event sequences help you infer their movement and actions.

\vspace{0.3cm}
\textbf{[Objective]} \\
Given the layout of the house and the description of the sensors, you need to analyse step by step the input sensor event sequences to determine their actions and tracks with the time.

\vspace{0.3cm}
\textbf{[House Layout]} \\
Start at the Hall (bottom-left). The Hall has two openings only: north into the Medicine Area, and east into the Kitchen. There is no direct Hall access to the Living Room, Dining Room, or Office. \\
From the Hall $\rightarrow$ Medicine Area $\rightarrow$ Living Room (east from Medicine Area) $\rightarrow$ Office (open passage to the east). \\
From the Hall $\rightarrow$ Kitchen $\rightarrow$ Dining Room (fully open to the east).

Overall arrangement by bands: \\
top row: Medicine Area (left), Living Room (center), Office (right); \\
bottom row: Hall (left), Kitchen (center), Dining Room (right). No other direct connections exist. \\
Each room contains specific facilities and sensors. The Hall serves as the entrance and passage area, with no sensors. The Medicine Area includes a medicine cabinet equipped with a medicine magnetic sensor (R6) to detect door openings. The Living Room contains 4 couches, each with a pressure mat (P3, P7, P8, P9). Face to the couches, there is a television connected to a smart plug (E2). Next to the living room, there is an office with a dest and an office chair (P2). The Kitchen, located in the center, includes several sensors: a cooker smart plug (E1) for the stove, a magnetic sensor (R5) for the pot cabinet, a magnetic sensor (R7) for the refrigerator door, a magnetic sensor (R1) for the pantry, and a magnetic sensor (R2) for the cutlery drawer. The Dining Room features a dining table with four dining chairs (P1, P4, P5, P6) and is designed as an open dining space connected to the Kitchen.

\vspace{0.3cm}
\textbf{[Sensor Description]} \\
SENSOR EXPLANATIONS (In column order)

\begin{tabbing}
\hspace{2cm} \= \hspace{3cm} \= \kill %
\textbf{Sensor ID} \> \textbf{Sensor Type} \> \textbf{Description} \\[3pt]
R1 \> Magnetic \> pantry \\
R2 \> Magnetic \> cutlery drawer \\
R5 \> Magnetic \> pots drawer \\
R6 \> Magnetic \> medicines cabinet \\
R7 \> Magnetic \> fridge \\
E1 \> Smart Plug \> stove \\
E2 \> Smart Plug \> television \\
P1 \> Pressure Mat \> dining room chair \\
P2 \> Pressure Mat \> office chair \\
P3 \> Pressure Mat \> living room couch \\
P4 \> Pressure Mat \> dining room chair \\
P5 \> Pressure Mat \> dining room chair \\
P6 \> Pressure Mat \> dining room chair \\
P7 \> Pressure Mat \> living room couch \\
P8 \> Pressure Mat \> living room couch \\
P9 \> Pressure Mat \> living room couch
\end{tabbing}

\vspace{0.3cm}
\textbf{[Key Constraints]} \\
When processing each sensor event step by step, you must carefully consider location constraints:
\begin{itemize}[leftmargin=*, label=-, noitemsep, topsep=0pt]
    \item Always check each user's recent activities to determine their current location status
    \item Consider time sequence and movement feasibility when assigning events
    \item If a user is already at an exclusive location (pressure mat, bathroom, etc.), they cannot be at a different exclusive location simultaneously
\end{itemize}

\vspace{0.3cm}
\textbf{[Task]} \\
Based on the historical activity patterns of each user and the current sensor event, predict which user(s) triggered the current event. Output user indices (1-indexed) as a multi-label prediction.

\end{tcolorbox}

\subsection{Inference}
At deployment, the LLM policy is decoded with standard greedy autoregressive generation: at each event step $t$, the prompt $\mathcal{X}_t$ is fed to the fine-tuned model, which emits a JSON response that is parsed into the assignment action $a_t$ and used to update the state to $s_{t+1}$. Unlike the multi-label policy of Section~\ref{sec: bc}, the LLM does not expose a closed-form per-slot probability vector $\bigl(p_t^{(1)}, \ldots, p_t^{(K)}\bigr)$; recovering the joint probability of an arbitrary action $a_t$ would require evaluating the likelihood of every candidate JSON string under the model, which scales poorly with the number of residents. We therefore do not apply the trajectory-level beam search of Section~\ref{sec: beam_search} to the LLM policy in this work, and treat the comparison between the two policies under their respective native decoders as a faithful reflection of how each is intended to be deployed.

\section{Experimental Evaluation}
\label{sect:experimentalevaluation}
\subsection{Datasets} \label{sec: datasets}
Among the mainstream multi-resident ambient sensor datasets are CASAS~\cite{singla2010recognizing}, ARAS~\cite{alemdar2013aras}, MARBLE~\cite{arrotta2021marble}, and the recently introduced MuRAL dataset~\cite{chen2025mural}. Notably, the ARAS dataset does not include resident identity annotations for individual sensor events, making it unsuitable for evaluating the Data Association task. As a result, ARAS is excluded from our experimental evaluation. 

\subsubsection{CASAS ADLMR Dataset}
CASAS ADLMR~\cite{singla2010recognizing} is a two-person activity recognition dataset. It was collected over 26 sessions in a home equipped with 37 ambient sensors. These sensors are primarily ceiling-mounted PIR motion sensors, which are triggered by movement within approximately one meter, along with item sensors used to indicate whether objects such as medications remain in their original locations. For each session, a different pair of residents performed a series of 15 activities. Each event triggered by an sensor is further annotated with one or two corresponding resident IDs and their activities. Finally, we note that CASAS is dominated by densely deployed ceiling-mounted motion sensors, whose activations are difficult to describe faithfully in natural language (e.g., lacking object- or location-specific semantics). Therefore, we do not run any LLM-based methods on CASAS in our experiments.

\subsubsection{MARBLE Dataset}
The MARBLE dataset~\cite{arrotta2021marble} consists of 76 sessions representing different scenarios, with each sessions potentially involving 1 to 4 residents. It was collected in an environment equipped with 16 ambient sensors, including 5 magnetic sensors, 2 smart plugs and 9 pressure mat. During data collection, residents were instructed to perform up to 13 different activities in a scripted sequence. Unlike the CASAS ADLMR datasets, the MARBLE dataset incorporates resident-specific wearable devices such as micro-localization, smartwatch, and phone usage sensors in addition to environmental sensors. However, since our method does not consider these devices, we use only the ambient sensor data in our experiments. In MARBLE, each sensor event is attributed to exactly one person, resulting in a one-to-one correspondence.

\subsubsection{MuRAL Dataset}
The MuRAL dataset~\cite{chen2025mural} is a multi-resident smart home dataset comprising 21 sessions, each involving 2 to 4 participants and instrumented with 23 ambient sensors (5 smart plugs, 10 magnetic sensors, 6 PIR motion sensors) . Every sensor event in the dataset is annotated with the identities of the triggering residents, one of 27 predefined activity categories, and a natural language description. Compared to existing datasets such as CASAS and MARBLE, MuRAL offers distinct advantages in the data association task by capturing dynamic resident entry and exit, and by allowing unscripted, naturalistic behaviors. This design results in richer and more complex activity patterns, better reflecting real-world multi-resident scenarios.

\subsection{Evaluation Metrics}
\subsubsection{Resident Assignment Accuracy (Acc)}

Prior evaluations of data association often suffer from global bias and cascading error effects. Some studies~\cite{riboni2020unsupervised, chen2024towards} assess performance only indirectly via downstream activity recognition, while others~\cite{crandall2008attributing, crandall2010using, hsu2010strategies} evaluate identity assignments based on a global one-to-one mapping between predicted and ground truth labels in a small chunk. However, since our method can process the entire sequence of a session, our evaluation scope is significantly larger. Once this mapping is misaligned—especially in the early part of a session—all subsequent predictions may be incorrectly penalized, even if identity tracking later becomes accurate.

To mitigate this issue, we propose a \emph{window-based evaluation protocol} that enables localized assessment of identity consistency. The key idea is to decouple the prediction phase from the evaluation phase: we first perform data association over the entire session to produce a complete sequence of predicted identity labels. We then slide a fixed-length window (by default, the window size $w$ is 16, the stride is 1) across the session and assess identity tracking consistency within each window.

In each window, we extract the resident labels from both the ground truth and predicted sequences, and assign them symbolic codes (A, B, C, ...) based on their order of first appearance. This yields a one-to-one mapping between raw labels and symbolic identifiers \emph{within} the window. We then relabel both the ground truth and prediction sequences using this mapping, and compute a per-event consistency score. The score for the window is the average consistency across all steps, and the final tracking score is the mean over all windows.

This evaluation strategy preserves temporal context from the full-session prediction, while preventing early errors from dominating the accuracy measurement. It provides a robust means to analyze \emph{cascading error}, as performance degradation under increasing window sizes reflects how well a method maintains identity continuity over time.

\subsubsection{Correct Run Length (CRL)}

To complement the accuracy metric, we introduce Correct Run Length (CRL) as an additional evaluation measure.
CRL directly reflects the quality of resident tracking by quantifying how long a model can continuously maintain correct identity assignments before the first error occurs.
Specifically, for each sliding window and for each ground-truth resident, we count the length of the initial consecutive segment of correct predictions until the first mistake.
These lengths are then averaged across all residents and all windows to produce the CRL score.
A larger CRL indicates that the model sustains more stable and coherent tracking of residents.

To contextualize CRL, we also report, for reference, the \emph{average number of events per resident}.
This auxiliary value allows readers to interpret CRL in terms of the typical sequence length over which the tracking is evaluated, but it is not used as a primary metric.

\subsubsection{BLEU}\label{sec: bleu}
We adopted the Bilingual Evaluation Understudy (BLEU)~\cite{papineni2002bleu} to assess the consistency of the seperated sequence. For any mixed sequence, we connect the separated sequence generated by the model using a delimiter (an <EOS> token) to form a candidate sequence. Similarly, we construct a reference sequence using the ground truth in the same way. Given a candidate sequence and a reference sequence, the BLEU metric considers the precision of $N$-grams, which is the proportion of $N$-gram phrases in the candidate sequence that appear in the reference sequence. This evaluation method focuses more on assessing whether the results of the model's separation are coherent and consistent. In our subsequent experiments, we divide the mixed sequence into segments of length 30 and evaluate the BLEU score for each segment. Finally, we compute the average BLEU score across all segments.

\subsubsection{Resident Count Error (RCE)}
Since BLEU and Resdient Assignment Accuracy cannot directly reflect the model's real-time assessment of the number of residents, we designed a dedicated error metric for the number of residents in order to systematically evaluated the estimation of the number of individuals in the results. This metric is introduced for the first time in related research, as previous methods lacked the ability to dynamically evaluate the number of residents. To compute this metric, we divide the mixed sequence into segments of 30 events. We chose 30 as the segment length which is sufficient to involve multiple residents while still keeping the local representation. For each segment, we count how many different residents the model assigns the \( L \) events to, obtaining a predicted value \( \hat{K} \). This predicted value is then compared with the ground truth \( K \), and the absolute error for each segment is calculated. Finally, the average of these absolute errors across all segments is computed.

The error metric can be expressed as:
\[
\text{RCE} = \frac{1}{S} \sum_{i=1}^{S} \left| K_i - \hat{K}_i \right|
\]
where \( S \) is the total number of segments, \( \hat{K}_i \) is the predicted number of residents for the \( i \)-th segment, and \( K_i \) is the ground truth number of residents for the \( i \)-th segment.

\subsection{Experimental Setup}

We conduct experiments on the three benchmark datasets described above (CASAS~\cite{singla2010recognizing}, MARBLE~\cite{arrotta2021marble}, and MuRAL~\cite{chen2025mural}) to compare our proposed method with sMRT~\cite{wang2020smrt}, LAHAR~\cite{chen2024towards} which is powered by GPT-4o, and the naive ``All-to-One'' baseline which assumes that all events belong to a single resident. To conduct these experiments, we randomly split the sessions of each datasets into training, validation, and test sets with a ratio of 7:1:2. For our proposed methods, we refer to the Transformer-only policy trained from scratch via Behavior Cloning as \textbf{BC}, and the approach that uses an LLM as the policy as \textbf{LLM+BC}. 

\subsubsection{Setting of BC}
The BC method utilizes a Transformer encoder, consisting of $3$ layers with a model dimension ($d_{\text{model}}$) of $128$, $8$ attention heads, and a feed-forward dimension of $512$. Training is performed with a batch size of $512$, a learning rate of $10^{-4}$, and a dropout rate of $0.2$. These model architectures and training hyperparameters were chosen based on preliminary trials on the validation set, without extensive hyperparameter tuning, to demonstrate the robustness and generalizability of the proposed method. However, to accommodate the varying complexity and density of the datasets, we adjusted the training duration and the resident history window size ($L$) for each dataset. Specifically, the MARBLE dataset was trained for $200$ epochs with a history window of $L=11$. In contrast, both the CASAS and MuRAL datasets were trained for $100$ epochs, with history windows set to $L=9$ and $L=1$, respectively. These hyperparameters were obtained via grid search on the validation set, and the search procedure and its impact will be presented in the following sections.
As for inference, we adopt greedy decoding by default, corresponding to a beam size of $1$, for a fairer comparison with prior work as well as with the LLM+BC variant under its native decoding setup, and apply a fixed multi-label decision threshold of $\tau = 0.5$ across all datasets. These settings are used as standard defaults without further tuning, in order to avoid dataset-specific optimization and to ensure fair and reproducible evaluation.

\subsubsection{Setting of LLM+BC}
We employ the Qwen3-4B-Instruct-2507 as the foundation model for all experiments. 
To enable parameter-efficient fine-tuning, we adopt Low-Rank Adaptation (LoRA)~\cite{hu2022lora}.
The LoRA configuration consists of a rank $r=16$, a scaling factor $\alpha=32$, and a dropout rate of $0.05$. This adaptation is applied to all linear projections within both the attention mechanisms and feed-forward networks. Training is performed using Bfloat16 (bf16) precision to ensure computational efficiency. All experiments were conducted using the Hugging Face \texttt{Transformers} Trainer framework and the Hugging Face \texttt{PEFT} library for the LoRA implementation. The computational setup consisting of two NVIDIA A100 (80GB) GPUs. To leverage hardware parallelism, we employed the \texttt{Accelerate} library for Distribued Data Parallel (DDP) training. The model was fine-tuned for $1$ epoch with a maximum sequence length of $1024$ tokens given the length of the prompt is about $900$ tokens. We set the per-device batch size to $6$ with gradient accumulation steps of $4$, resulting in an effective global batch size of $48$ ($6 \times 2 \times 4$). Optimization was performed using a cosine learning rate scheduler with a peak learning rate of $2 \times 10^{-4}$ and a warmup ratio of $0.1$, regularized by a weight decay of $0.01$. For inference, we configure the generation process to produce a maximum of $50$ new tokens. We align our history context lengths $L$ with those found in BC method, using a window of $1$ for MuRAL and $11$ for MARBLE.

\subsection{Qualitative Examples}\label{sec:qualitative_examples}
\begin{table*}[htbp]
\centering
\tiny
\setlength{\tabcolsep}{4pt}
\renewcommand{\arraystretch}{1.5}
\caption{Qualitative example of the resident assignment performed by our LLM+BC method on 33 sequential sensor events from Session 12 of the MuRAL dataset. Row colors indicate ground truth residents: green (A), blue (B), and orange (C). Red rows represent incorrect predictions.}
\label{tab:qualitative-full}
\begin{tabular}{rcccccc}
\toprule
\textbf{Step} & \textbf{Time} & \textbf{Event} & \textbf{Facility} & \textbf{Predicted ID} & \textbf{GT ID} & \textbf{Correct} \\
\midrule
\rowcolor{lightgreen}
0  & 19:04:24 & OPEN  & Main Door & A, B & A, B & \checkmark \\
\rowcolor{lightgreen}
1  & 19:04:30 & CLOSE & Main Door & A, B & A, B & \checkmark \\
\rowcolor{lightgreen}
2  & 19:04:35 & ON    & Living Room Couch & A & A & \checkmark \\
\rowcolor{lightblue}
3  & 19:04:36 & OPEN  & Bedroom 1 Door & B & B & \checkmark \\
\rowcolor{lightgreen}
4  & 19:04:41 & OFF   & Living Room Couch & A & A & \checkmark \\
\rowcolor{lightpink}
5  & 19:04:42 & ON    & Bathroom Global & B & A & \ding{55} \\
\rowcolor{lightpink}
6  & 19:04:42 & CLOSE & Bedroom 1 Door & A & B & - \\
\rowcolor{lightpink}
7  & 19:04:46 & ON    & Living Room Couch & A & B & - \\
\rowcolor{lightpink}
8  & 19:04:49 & CLOSE & Toilet Contact & B & A & - \\
\rowcolor{lightpink}
9  & 19:04:52 & OFF   & Living Room Couch & A & B & - \\
\rowcolor{lightpink}
10 & 19:05:00 & OFF   & Bathroom Global & B & A & - \\
\rowcolor{lightpink}
11 & 19:05:26 & ON    & Bathroom Global & B & A & - \\
\rowcolor{lightpink}
12 & 19:05:37 & OPEN  & Toilet Contact & B & A & - \\
\rowcolor{lightpink}
13 & 19:05:44 & ON    & Bathroom Sink & B & A & - \\
\rowcolor{lightpink}
14 & 19:05:52 & OFF   & Bathroom Sink & B & A & - \\
\rowcolor{lightpink}
15 & 19:06:04 & OFF   & Bathroom Global & B & A & - \\
\rowcolor{lightpink}
16 & 19:06:13 & OPEN  & Fridge Door & B & A & - \\
\rowcolor{lightpink}
17 & 19:06:14 & CLOSE & Fridge Door & B & A & - \\
\rowcolor{lightpink}
18 & 19:07:58 & OPEN  & Dining Chair 2 & B & A & - \\
\rowcolor{lightpink}
19 & 19:08:06 & ON    & Living Room Couch & A & B & - \\
\rowcolor{lightpink}
20 & 19:08:10 & OFF   & Living Room Couch & A & B & - \\
\rowcolor{lightpink}
21 & 19:08:16 & OPEN  & Dining Chair 3 & A & B & - \\
\rowcolor{lightorange}
22 & 19:11:49 & OPEN  & Main Door & C & C & \checkmark \\
\rowcolor{lightorange}
23 & 19:11:53 & CLOSE & Main Door & C & C & \checkmark \\
\rowcolor{lightorange}
24 & 19:12:06 & OPEN  & Bedroom 2 Door & C & C & \checkmark \\
\rowcolor{lightpink}
25 & 19:12:09 & CLOSE & Dining Chair 2 & B & A & - \\
\rowcolor{lightpink}
26 & 19:12:10 & CLOSE & Dining Chair 3 & A & B & - \\
\rowcolor{lightorange}
27 & 19:12:12 & CLOSE & Bedroom 2 Door & C & C & \checkmark \\
\rowcolor{lightpink}
28 & 19:12:21 & ON    & Kitchen Sink & B & A & - \\
\rowcolor{lightblue}
29 & 19:12:24 & ON    & Kitchen Counter & B & B & \checkmark \\
\rowcolor{lightpink}
30 & 19:12:43 & OFF   & Kitchen Sink & B & A & \ding{55} \\
\rowcolor{lightpink}
31 & 19:12:51 & ON    & Stove & B & A & - \\
\rowcolor{lightorange}
32 & 19:13:07 & OPEN  & Bedroom 2 Door & C & C & \checkmark \\
\bottomrule
\end{tabular}
\end{table*}

Table~\ref{tab:qualitative-full} illustrates the resident assignment performance of our proposed LLM+BC method, using a sequential excerpt of 33 sensor events from Session 12 of the MuRAL dataset.

While the system correctly initializes tracks for Residents A and B upon their simultaneous entry (Steps 0--3), a critical identity switch occurs at Step 5. We hypothesize that this early failure stems from the model's rigid interpretation of temporal constraints. As shown in the log, the Living Room Couch turns \texttt{OFF} at 19:04:41 (Step 4), and the Bathroom turns \texttt{ON} merely one second later at 19:04:42 (Step 5). The LLM likely inferred that it was physically infeasible for Resident A to traverse from the living room to the bathroom within such a negligible time window. Consequently, to satisfy this apparent physical constraint, the model assigned the bathroom activity to Resident B, leading to a persistent identity swap for the subsequent events. In reality, the sensor timestamps did not perfectly align with physical movements; Resident A had placed a bag on the couch at Step 2 and immediately departed. The \texttt{OFF} event at Step 4 was merely a delayed deactivation caused by the sensor's timeout mechanism, which the model failed to account for.

A second challenge arises during the interaction in the kitchen area (Steps 25--31), illustrating the difficulty of tracking concurrent activities in shared zones. At Steps 25 and 26, distinct sensors on Dining Chairs 2 and 3 trigger \texttt{CLOSE} events almost simultaneously, indicating that two individuals stood up to move elsewhere. However, the model fails to register this as a dual transition. When activities subsequently occur in the kitchen (Sink, Counter, and Stove at Steps 28--31), the model attributes them solely to a single resident (Predicted ID: B). This suggests that the model overlooked the cue provided by the simultaneous chair events, failing to realize that both residents had entered the kitchen. Consequently, it collapses the multi-resident scenario into a single-resident track, highlighting the complexity of distinguishing identities when residents occupy the same confined functional area.

Despite these challenges with Residents A and B, the system demonstrates robustness in handling cardinality changes. When Resident C enters the facility at Step 22, the model successfully distinguishes this new individual from the existing entangled tracks. The events associated with Resident C (Steps 22--32) are tracked with perfect accuracy, indicating that the reasoning engine remains effective for distinct, unambiguous trajectories.

\subsection{Quantitative Comparison against Baselines}

Table~\ref{tab:bc_comparison_sota} presents the performance comparison between our proposed methods and the baselines. A decisive advantage of our approaches, BC and LLM+BC, is first observed in the Resident Count Error (RCE), where they demonstrate a substantial improvement over the All-to-One and sMRT~\cite{wang2020smrt} baselines across all datasets. This contrast is most distinct on the MARBLE dataset: while the sMRT filter suffers from error covariance divergence due to sparse sensor data, resulting in a catastrophic RCE of 15.10, our methods maintain exceptional stability with an RCE as low as 0.136. This highlights the capability of our MDP-based models to dynamically estimate the number of active residents, whereas traditional filtering algorithms tend to erroneously assign events to independent identities under high uncertainty.

\begin{table}[t!]
\centering
\caption{Comparison of Data Association performances of baseline approaches and our approach on 3 datasets. Acc and CRL are calculated using a window size $w=16$.}
\label{tab:bc_comparison_sota}
\resizebox{0.9\textwidth}{!}{
\begin{tabular}{l|cccc|cccc|cccc}
\hline
Dataset       & \multicolumn{4}{c|}{CASAS~\cite{singla2010recognizing}} & \multicolumn{4}{c|}{MARBLE~\cite{arrotta2021marble}} & \multicolumn{4}{c}{MuRAL~\cite{chen2025mural}} \\ \hline
Metric        & Acc                & CRL               & BLEU              & RCE               & Acc               & CRL              & BLEU             & RCE              & Acc             & CRL             & BLEU            & RCE            \\ \hline
All-to-One        & 0.580                  & 5.372                 & 0.565                 & 0.995                 & 0.729             & 8.606            & 0.707            & 0.789            & 0.551 & 4.489           & 0.510           & 1.593          \\
sMRT~\cite{wang2020smrt}       & 0.561                  & 3.051                 & 0.530             & 1.257             & 0.119                 & 0.641                & 0.079            & 15.10            & 0.529               & 3.321               & 0.497           & 1.560          \\
LAHAR~\cite{chen2024towards}       & /                  & /                 & /             & /             & 0.730                 & 8.231                & 0.747            & 0.383            & 0.616               & 4.375               & 0.736           & 0.571          \\
BC & \textbf{0.757}     & \textbf{6.234}    & \textbf{0.727}    & \textbf{0.021}    & 0.792    & 8.710   & 0.797   & 0.136   & 0.728  & 5.476  & 0.774  & 0.437 \\ 
LLM+BC & /     &  /  & /    & /   & \textbf{0.824}             & \textbf{8.783}   & \textbf{0.841}            & \textbf{0.136}   & \textbf{0.773}  & \textbf{6.034}           & \textbf{0.799}           & \textbf{0.261}          \\ \hline
\end{tabular}}
\end{table}

Significantly, our approach outperforms the recent LLM-driven baseline, LAHAR~\cite{chen2024towards}, despite a massive disparity in foundation model scale. While LAHAR relies on the state-of-the-art commercial model \textbf{GPT-4o}, our LLM+BC framework utilizes the compact, open-source \textbf{qwen3-4B}. Notwithstanding this resource constraint, our method achieves superior performance across all metrics. This advantage is particularly pronounced on the complex MuRAL dataset, where LLM+BC surpasses LAHAR by a wide margin in Accuracy (0.773 vs. 0.616) and reduces the RCE by over 50\% (0.261 vs. 0.571). These results suggest that our structured MDP formulation provides more effective grounding for sensor data than relying solely on the generalized reasoning capabilities of extremely large language models.

Comparing our two proposed methods reveals the specific contribution of the semantic module. While pure BC is effective in simpler settings, the LLM+BC framework provides a critical advantage in complex scenarios like MuRAL. This is best reflected in the Correct Run Length (CRL) and BLEU metrics. On MuRAL, incorporating the qwen3-4B model improves the CRL from 5.476 to 6.034. Since CRL measures the length of consecutive correct predictions, this increase signifies that the semantic reasoning of the LLM significantly enhances tracking stability, sustaining correct identity assignments for longer durations. Similarly, the improvement in BLEU score (from 0.774 to 0.799) indicates that the sequences generated by LLM+BC possess higher coherence and better align with the N-gram structure of ground truth behaviors.

\subsection{Impact of Policy Training}
\begin{figure}[t!]
    \centering
    \includegraphics[width=\linewidth]{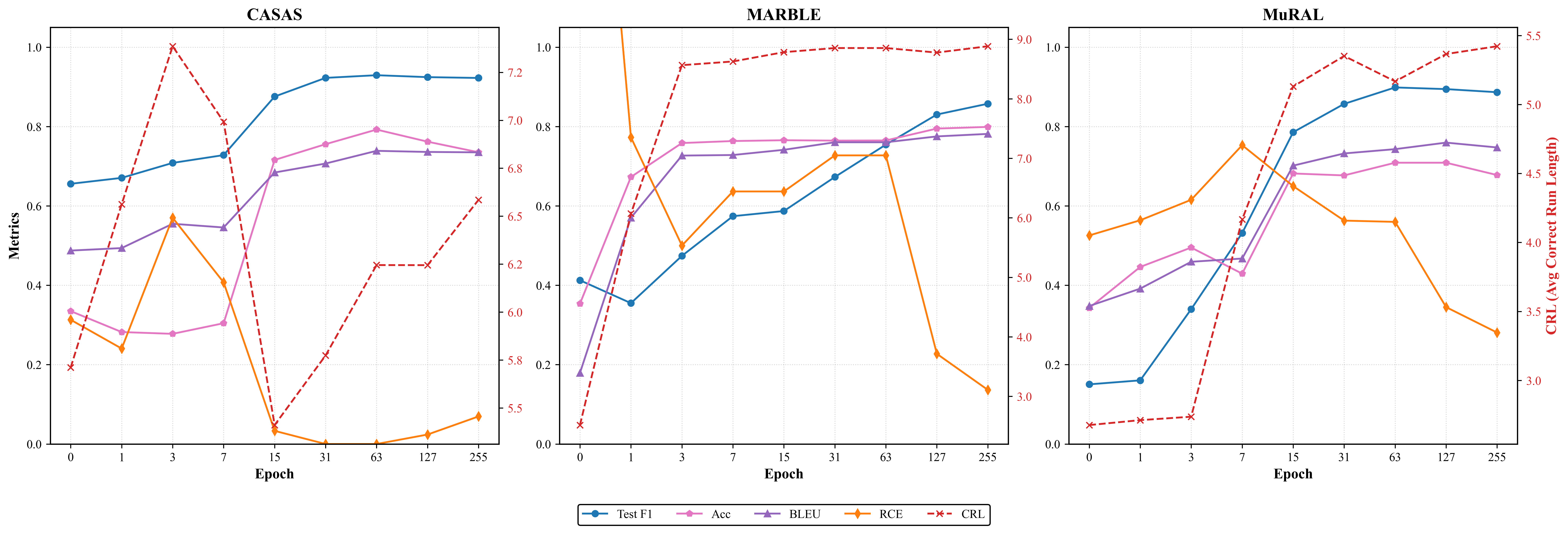}
    \caption{Performance evolution of the policy network across training epochs on all datasets.}
    \label{fig:bc_epoch_performances}
\end{figure}

Our proposed MDP-based framework relies critically on the learned policy model to make sequential resident-assignment decisions. To validate the importance and effectiveness of this policy component, we study how BC policy models at different training stages affect the final Data Association performance. Figure~\ref{fig:bc_epoch_performances} plots the test-set F1 score of the policy model together with the Data Association evaluation metrics (Acc, BLEU, RCE, and CRL) obtained during inference for different training epochs on the three datasets.

Across all datasets, we observe that the policy network’s classification performance generally improves as training progresses: the test-set F1 score exceeds 0.8 in all cases and approaches 0.9 on both CASAS and MuRAL. This trend suggests that the mapping from multi-resident contextual event representations to resident-assignment actions is learnable under the current dataset scale and model capacity. Moreover, as the test-set F1 score increases, the downstream Data Association metrics (Acc, BLEU, RCE, and CRL) improve in tandem, indicating a strong linkage between per-step policy accuracy and sequence-level association quality in the MDP inference process. Overall, these results support both our modeling formulation and the effectiveness of the BC-based training paradigm.

Beyond this overall trend, each dataset exhibits characteristic learning dynamics. On CASAS, policy F1 and BLEU generally increase with training, whereas RCE and CRL show noticeable non-monotonic fluctuations before epoch 15. We attribute these fluctuations to unstable early-stage policies (e.g., collapsing most events to a single resident), which can distort both resident-count estimates and run-length statistics. MARBLE shows faster early saturation in Acc and BLEU, likely because a large fraction of its sessions involve only one resident, making the underlying classification problem comparatively simpler. In contrast, CRL improves more gradually along with test-set F1, suggesting that sustaining longer streaks of correct identity assignments requires a more precisely trained policy, even when average accuracy has largely converged.

Overall, these results further confirm that improving the policy network translates into better sequence-level Data Association performance.

\subsection{Impact of Error Accumulation}\label{sec:error_accumulation}

\begin{figure}[t!]
    \centering
    \subcaptionbox{CASAS\label{fig:bc_win_size_casas}}{%
        \includegraphics[width=0.32\linewidth]{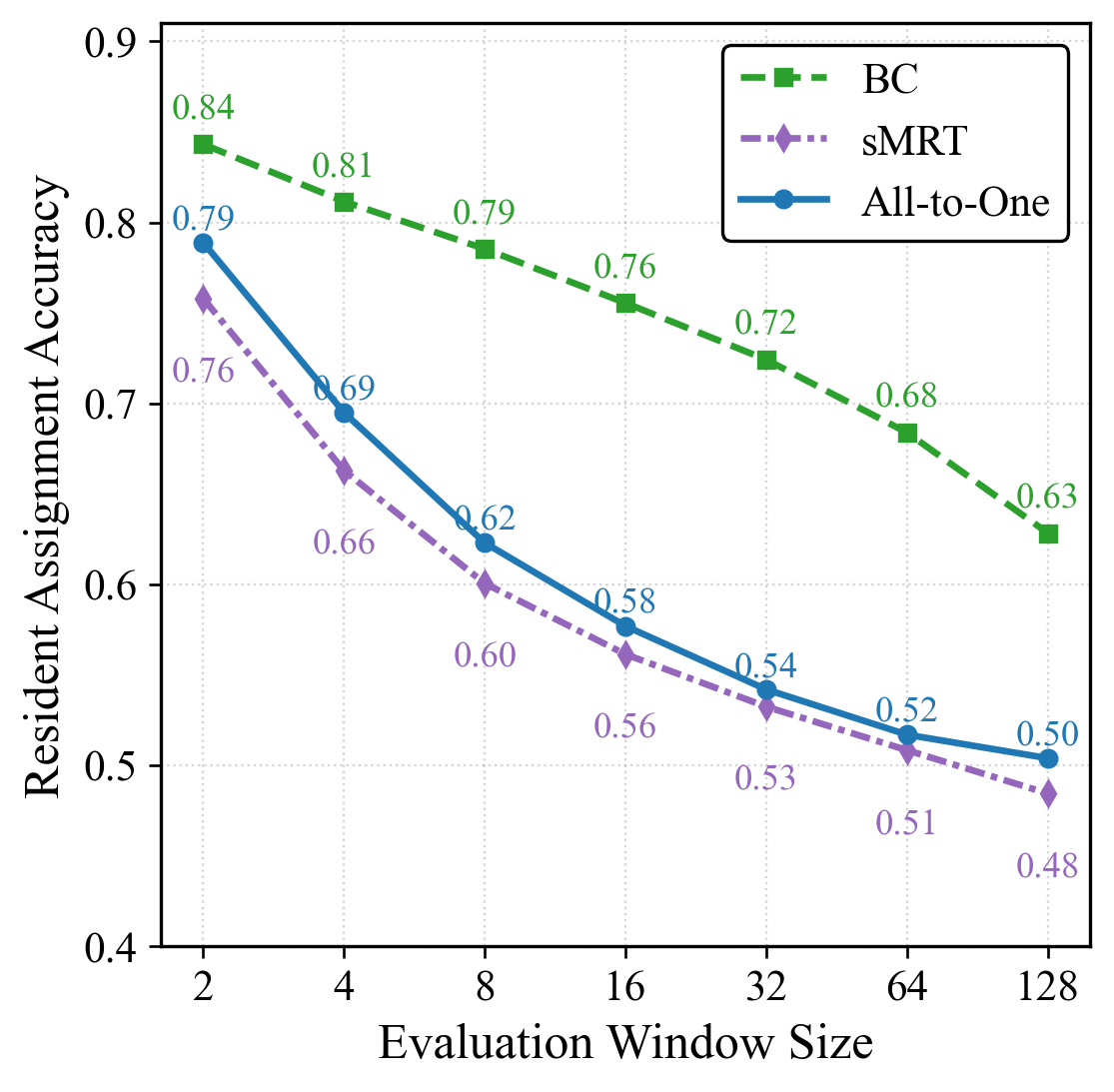}}
    \subcaptionbox{MARBLE\label{fig:bc_win_size_marble}}{%
        \includegraphics[width=0.32\linewidth]{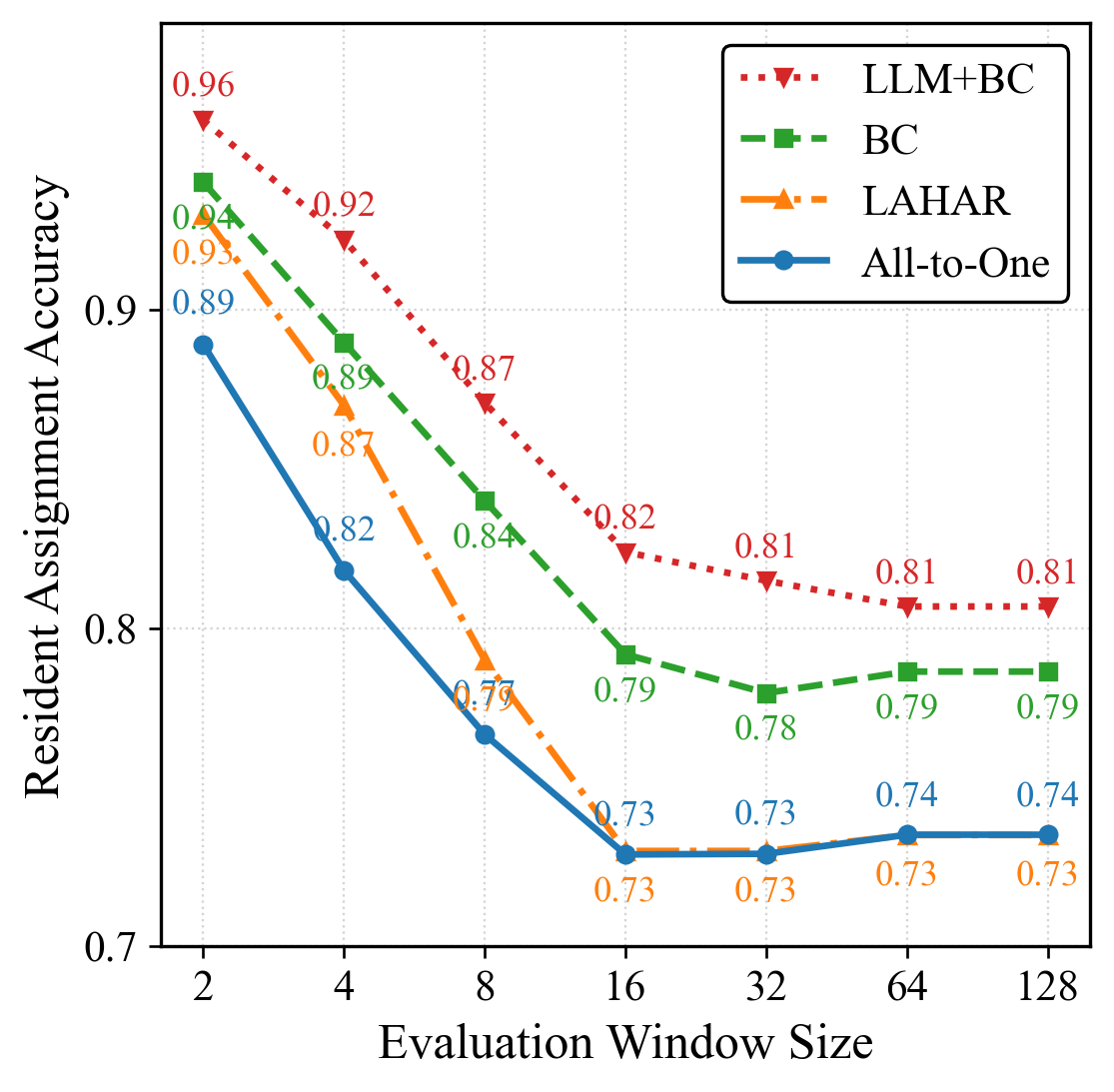}}
    \subcaptionbox{MuRAL\label{fig:bc_win_size_mural}}{%
        \includegraphics[width=0.32\linewidth]{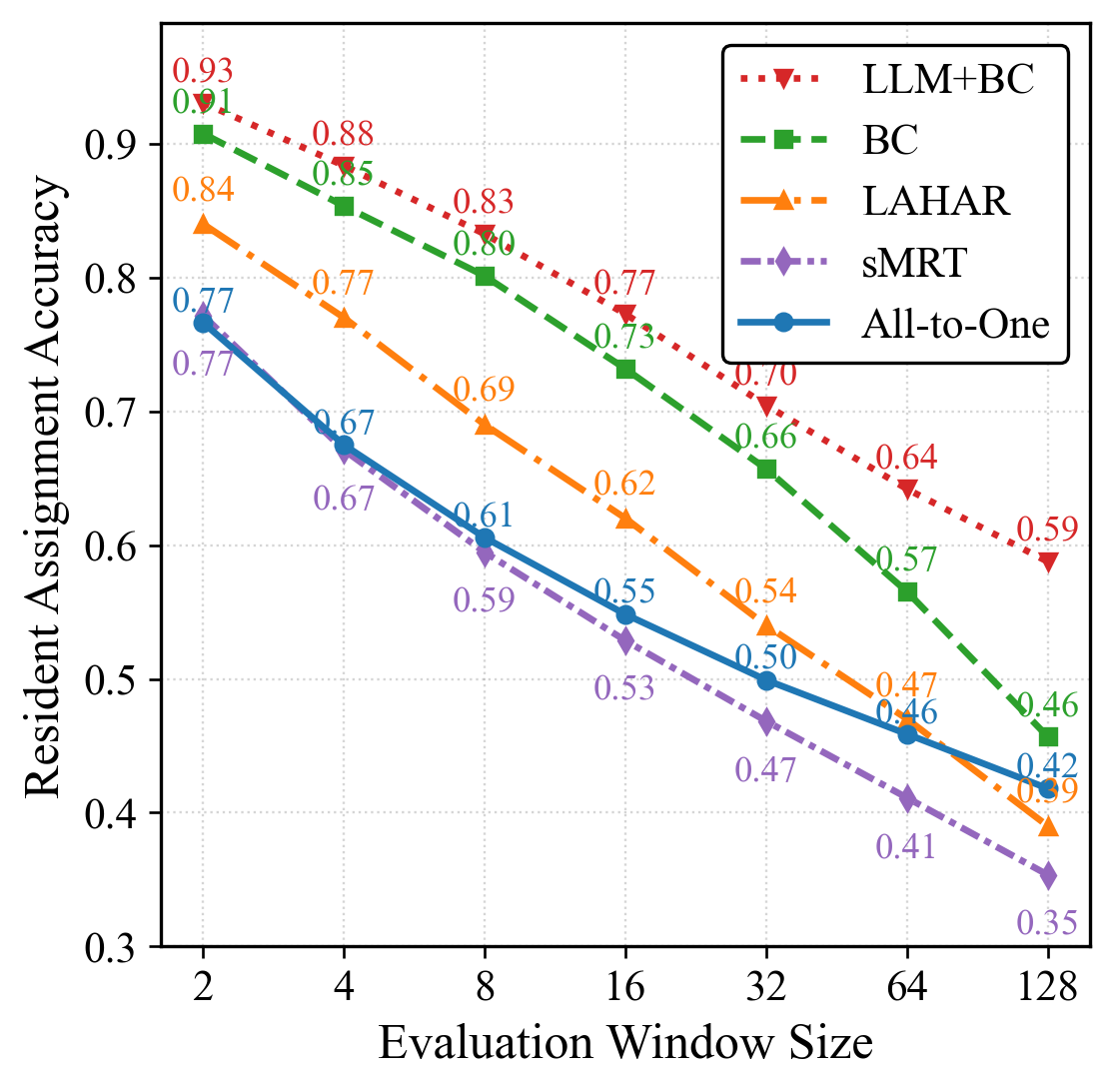}}
    \caption{Resident assignment accuracy as a function of the evaluation window size (log$_2$ scale) 
    on the CASAS, MARBLE, and MuRAL datasets.}
    \label{fig:bc_acc_vs_w}
\end{figure}

Data Association is inherently a sequential decision-making problem: a model assigns an identity to each incoming event conditioned on its current belief about all residents. As a result, mistakes are difficult to fully avoid and can compound over time. Once an early event is assigned to an incorrect resident, the resident states maintained by the model may deviate from the true states, and subsequent decisions are made under this shifted input distribution, leading to cascading errors.

To assess robustness to this error accumulation, we plot Acc as a function of the evaluation window size $w$ (log$_2$ scale) in Figure~\ref{fig:bc_acc_vs_w}. Larger $w$ corresponds to a longer temporal horizon over which the predicted identity pattern must remain consistent. Therefore, a faster degradation of Acc as $w$ increases indicates higher sensitivity to compounding errors. Notably, on MARBLE the filtering-based baseline sMRT fails completely under long-horizon evaluation: its Acc drops from 0.69 at $w=2$ to 0.08 at $w=32$. For readability and to keep the scale of Figure~\ref{fig:bc_acc_vs_w} informative for the remaining methods, we omit the sMRT curve from the plot of MARBLE.

Across all three datasets, all methods exhibit cascading errors: Acc generally decreases as the evaluation window size $w$ increases. On MARBLE, the curves tend to saturate once $w$ exceeds 16, which is consistent with the fact that each MARBLE session is relatively short (26.21 events on average). Beyond this point, enlarging the window provides limited additional temporal context to evaluate.

Within the range $w\in[2,32]$, our BC and LLM+BC degrade more slowly than the baselines on all three datasets, indicating stronger robustness to error accumulation in the short-to-medium horizon. We also observe that BC on CASAS and MuRAL shows a noticeably steeper drop for $w>32$, suggesting that long-horizon performance can still be affected by compounding errors. On MuRAL, LLM+BC maintains a milder degradation trend than BC when $w>32$, indicating improved robustness under larger distribution shift. We conjecture that this benefit may stem from the LLM's semantic priors, which help stabilize resident-state estimation when the model's own prediction history becomes less reliable.

\subsection{Inference Benefit of the MDP Formulation}\label{sec: exp_beam_search}
\begin{table}[t]
\centering
\caption{Impact of beam-search inference on the Data Association performance of our method. "Greedy" denotes standard decoding with beam size $1$, and "BS $k$" denotes beam search with beam size $k$.}
\label{tab:search_algo}
\begin{tabular}{l|l|ccccc}
\hline
Dataset                                                                                        & Metric & Greedy & BS 2           & BS 4           & BS 8           & BS 12          \\ \hline
\multicolumn{1}{c|}{\multirow{4}{*}{CASAS~\cite{singla2010recognizing}}} & Acc    & 0.757  & 0.758          & \textbf{0.760} & 0.759          & 0.759          \\
\multicolumn{1}{c|}{}                                                                          & CRL    & 6.234  & 6.579          & \textbf{6.811} & 6.797          & 6.666          \\
\multicolumn{1}{c|}{}                                                                          & BLEU   & 0.727  & 0.736          & \textbf{0.743} & 0.742          & 0.738          \\
\multicolumn{1}{c|}{}                                                                          & RCE    & 0.021  & 0.011          & \textbf{0.005} & 0.005          & 0.010          \\ \hline
\multirow{4}{*}{MARBLE~\cite{arrotta2021marble}}                         & Acc    & 0.792  & \textbf{0.797} & 0.795          & 0.795          & 0.794          \\
                                                                                               & CRL    & 8.710  & \textbf{8.765} & 8.761          & 8.760          & 8.756          \\
                                                                                               & BLEU   & 0.797  & \textbf{0.809} & 0.809          & 0.808          & 0.801          \\
                                                                                               & RCE    & 0.136  & \textbf{0.113} & 0.119          & 0.120          & 0.123          \\ \hline
\multirow{4}{*}{MuRAL~\cite{chen2025mural}}                              & Acc    & 0.728  & 0.733          & 0.744          & \textbf{0.747} & 0.747          \\
                                                                                               & CRL    & 5.476  & 5.628          & 5.761          & 5.822          & \textbf{5.833} \\
                                                                                               & BLEU   & 0.774  & 0.779          & 0.782          & 0.787          & \textbf{0.788} \\
                                                                                               & RCE    & 0.437  & 0.421          & 0.376          & \textbf{0.312} & 0.313          \\ \hline
\end{tabular}
\end{table}

As discussed in Section~\ref{sec: beam_search}, one key advantage of formulating Data Association as an MDP, compared with earlier filtering-based approaches, is that it enables the use of trajectory-level inference strategies such as beam search to handle complex and ambiguous events at test time. To validate this hypothesis, we compared greedy decoding ($B{=}1$) with the beam-search using widths $B \in \{2, 4, 8, 12\}$, holding the trained BC policy $\pi_\theta$ fixed.

Table~\ref{tab:search_algo} reports the resulting Data Association metrics on all three datasets. Across all datasets, beam search consistently matches or improves upon greedy decoding under the same policy weights, with the clearest gains appearing in long-term metrics such as CRL and RCE. For example, CRL increases from $6.234$ to $6.811$ on CASAS and from $5.476$ to $5.822$ on MuRAL. By contrast, per-step accuracy remains relatively stable, which is expected because beam search re-ranks complete trajectories rather than isolated decisions. Performance generally saturates around $B{=}4$--$8$, with little additional benefit at $B{=}12$, suggesting that the relevant ambiguity is concentrated in only a small number of plausible trajectories. Overall, these results show that trajectory-level search provides a practical inference benefit without changing the policy or training data.

\begin{figure}[t!]
    \centering
    \subcaptionbox{CASAS\label{fig:bc_win_size_casas2}}{%
        \includegraphics[width=0.32\linewidth]{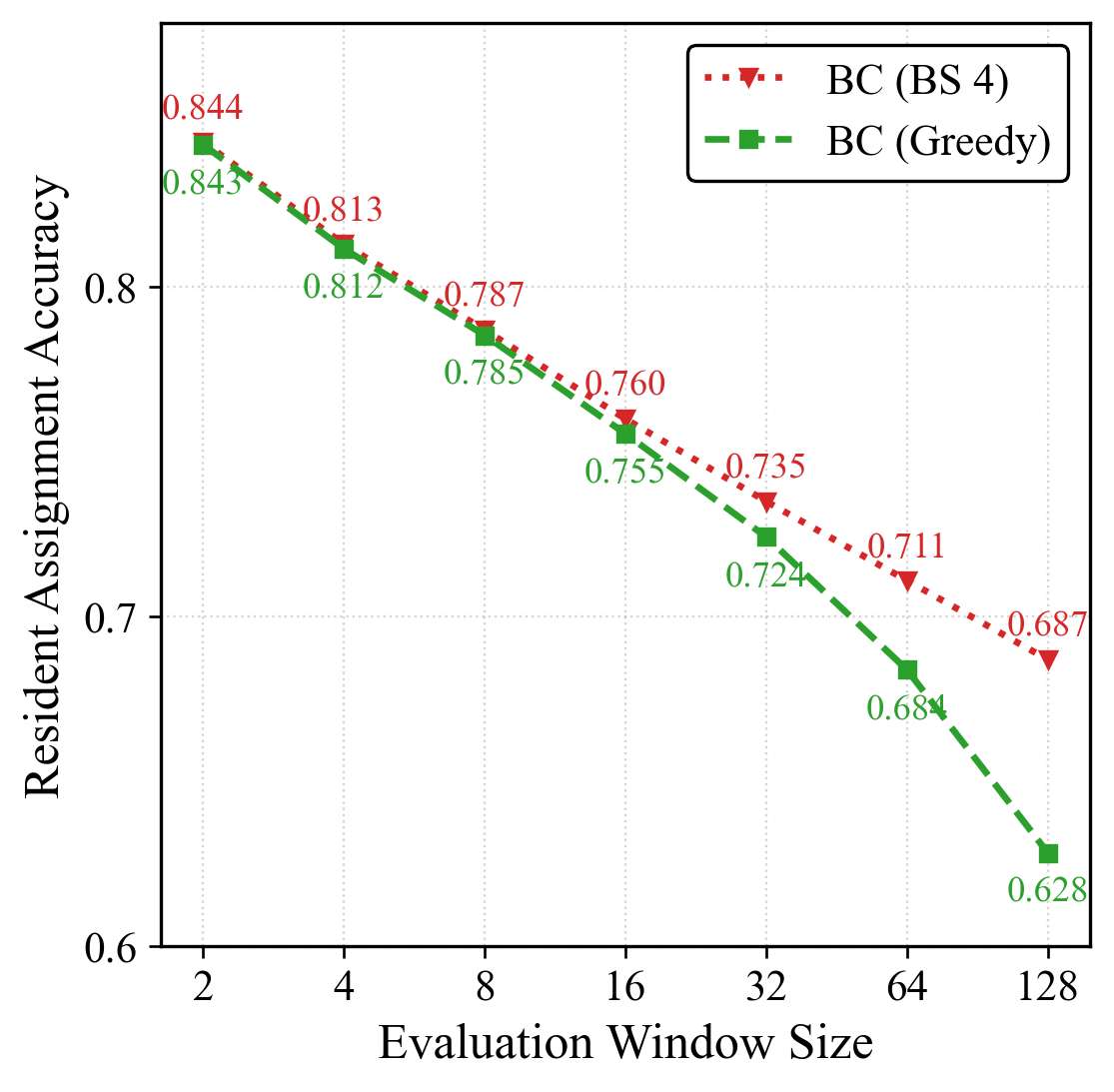}}
    \subcaptionbox{MARBLE\label{fig:bc_win_size_marble2}}{%
        \includegraphics[width=0.32\linewidth]{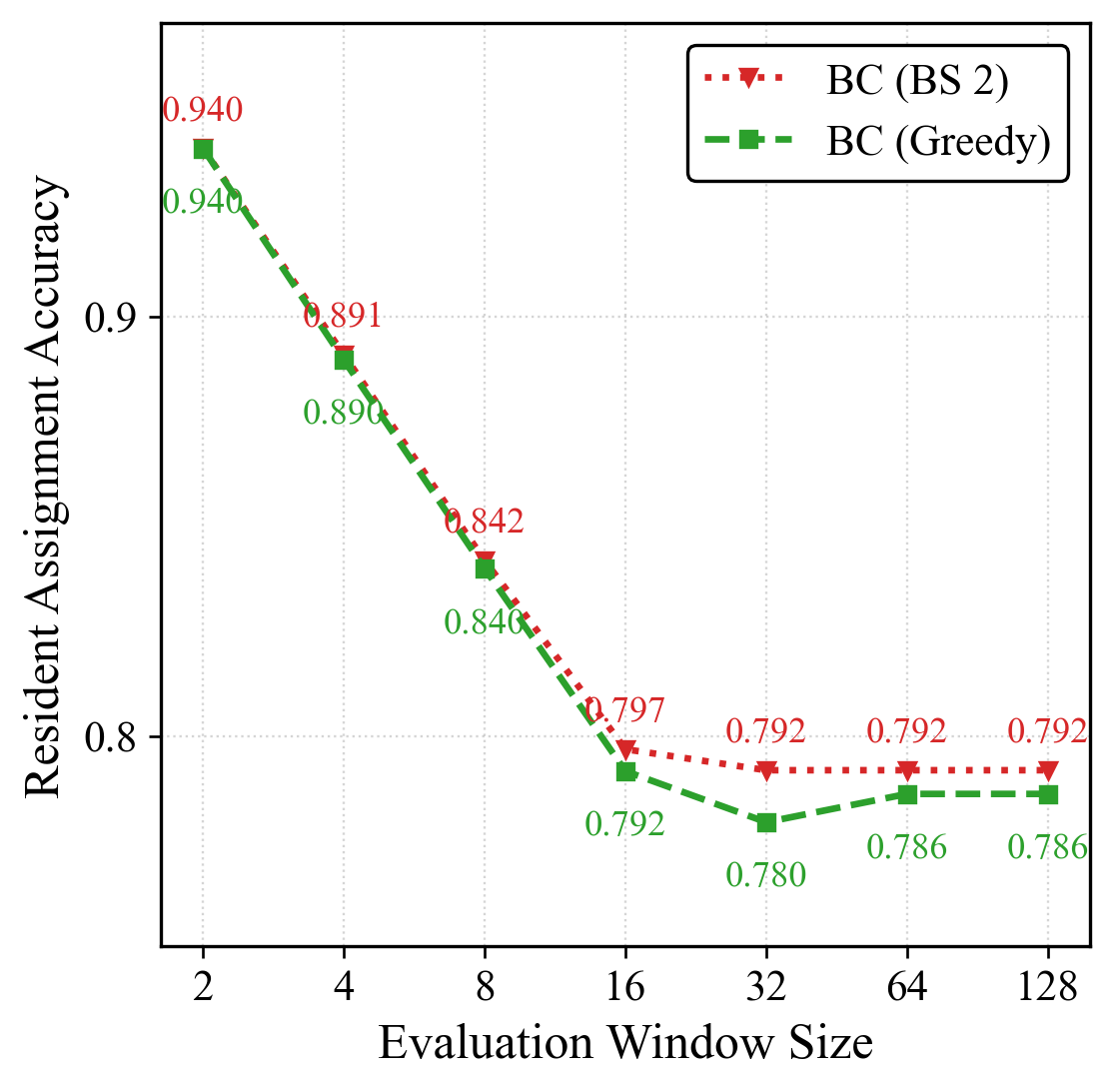}}
    \subcaptionbox{MuRAL\label{fig:bc_win_size_mural2}}{%
        \includegraphics[width=0.32\linewidth]{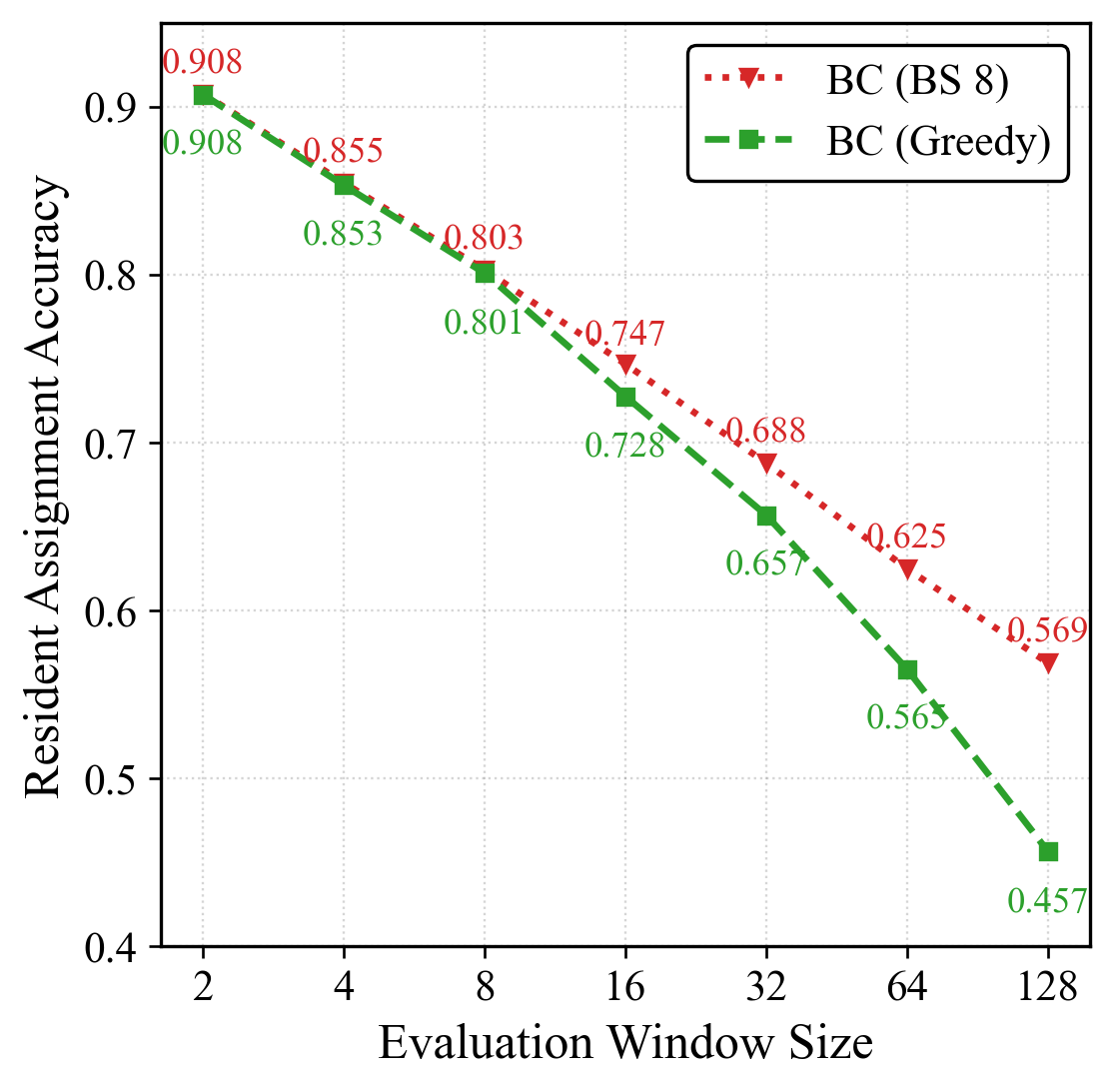}}
    \caption{Resident assignment accuracy as a function of the evaluation window size ($\log_2$ scale) under greedy decoding and beam-search decoding on the CASAS, MARBLE, and MuRAL datasets. A slower decline as the window size increases indicates better resistance to long-horizon error accumulation.}
    \label{fig:bc_acc_vs_bs}
\end{figure}

Figure~\ref{fig:bc_acc_vs_bs} further shows that the gap between beam search and greedy decoding widens as the evaluation window increases, indicating that beam search better mitigates error accumulation over longer horizons. This trend is most evident on MuRAL, where greedy accuracy drops from 0.908 to 0.457 when the window size reaches 128. By maintaining multiple candidate trajectories instead of committing to a single early decision, beam search is less vulnerable to transient ambiguities and therefore preserves stronger long-range consistency. These results further support the role of trajectory-level search in our MDP-based formulation.

\begin{table*}[t!]
\centering
\tiny
\setlength{\tabcolsep}{4pt}
\renewcommand{\arraystretch}{1.5}
\caption{Qualitative example illustrating how beam search resolves an early ambiguity that greedy decoding commits to. The table shows the first five sensor events from Session P20 of the CASAS ADLMR dataset, together with the ground-truth resident label, the prediction under greedy decoding, and the prediction under beam search (BS). Green prediction cells indicate correct assignments, red prediction cells indicate incorrect assignments, and the intensity of blue in the probability columns reflects the magnitude of the predicted assignment probability.}
\label{tab:beam_search_example}
\begin{tabular}{rcccccc}
\toprule
\textbf{Step} & \textbf{Event} & \textbf{GT ID} & \textbf{Greedy Pred.} & \textbf{Greedy Prob.} & \textbf{BS Pred.} & \textbf{BS Prob.} \\
\midrule
0  & M15 ON & A & \cellcolor{lightpink}A, B & \cellcolor{lightblue!44}0.438 & \cellcolor{lightgreen}A & \cellcolor{lightblue!43}0.426 \\
1  & M16 ON & A & \cellcolor{lightpink}A, B & \cellcolor{lightblue!93}0.928 & \cellcolor{lightgreen}A & \cellcolor{lightblue!92}0.916 \\
2  & D07 OPEN & A & \cellcolor{lightpink}A, B & \cellcolor{lightblue!99}0.987 & \cellcolor{lightgreen}A & \cellcolor{lightblue!99}0.993 \\
3  & M17 ON & A & \cellcolor{lightpink}A, B & \cellcolor{lightblue!99}0.989 & \cellcolor{lightgreen}A & \cellcolor{lightblue!99}0.985 \\
4  & M03 ON & B & \cellcolor{lightpink}A & \cellcolor{lightblue!31}0.311 & \cellcolor{lightgreen}B & \cellcolor{lightblue!88}0.875 \\
\bottomrule
\end{tabular}
\end{table*}

Table~\ref{tab:beam_search_example} provides a concrete example of how trajectory-level search resolves local ambiguity. At Step 0 (\texttt{M15 ON}), the event is ambiguous because it is the first observation in the session and no historical context is available. In addition, this sensor is statistically associated with situations in which two residents often co-occur. Accordingly, the policy predicts A with probability 0.865 and B with probability 0.507, while the remaining slots are close to zero. Under the multi-label factorization, this gives a probability of $0.865 \times 0.507 = 0.438$ for selecting both A and B, compared with $0.865 \times (1-0.507) = 0.426$ for selecting only A. The margin is therefore extremely small, indicating that the model has little confidence in the correct assignment of this event. Greedy decoding nevertheless forces a hard decision and commits to the joint assignment (\textit{A, B}), which is then propagated through Steps 1--3 around the \texttt{D07/M17} area, with confidence increasing to $0.989$. The consequence becomes clear at Step 4 (\texttt{M03 ON}), where the sensor \texttt{M03} is far from \texttt{M17}. Under the greedy trajectory, both A and B have already been tied to the \texttt{M17} region by the earlier joint assignment, so the new event is inconsistent with the current resident states. Greedy decoding still explains it using one of the existing labels and therefore makes a low-confidence misassignment to A. Specifically, the slot probabilities are 0.537 for A and 0.421 for B, so the probability of selecting only A is $0.537 \times (1-0.421) = 0.311$. Beam search instead preserves alternative hypotheses. Under the better hypothesis, resident B has not yet been instantiated, whereas resident A remains associated with the \texttt{M17} area and is therefore unlikely to trigger the distant \texttt{M03} sensor. Assigning the event to a newly created B thus becomes the most plausible explanation, and beam search assigns \texttt{M03} to B with high confidence ($0.875$). This example shows that beam search handles early ambiguity more effectively by delaying irreversible decisions until sufficient evidence becomes available.

\subsection{Assignments of Multi-Resident Events}~\label{sec:multi_events}

\begin{table}[t!]
\centering
\caption{Performance comparison of BC and LLM+BC in assigning events to the correct number of residents on CASAS and MuRAL.}
\label{tab:bc_multi_resident_event}

\begin{tabular}{l|l|c c|c c}
\hline
\multicolumn{1}{l|}{\#Residents} & Dataset & \multicolumn{2}{c|}{CASAS} & \multicolumn{2}{c}{MuRAL} \\ \cline{2-6} 
\multicolumn{1}{c|}{} & Method & LLM+BC & BC & LLM+BC & BC \\ \hline
\multirow{3}{*}{Multi} & Correct & / & 0.354 & 0.413 & 0.418 \\
 & Over-Assigned & / & 0.000 & 0.000 & 0.010 \\
 & Under-Assigned & / & 0.646 & 0.587 & 0.572 \\ \hline
Single & Over-Assigned & / & 0.081 & 0.008 & 0.028 \\ \hline
\end{tabular}
\end{table}

A multi-resident event refers to a single ambient sensor activation jointly triggered by multiple individuals, such as two co-located residents simultaneously activating the same motion sensor. Such events constitute 14.4\% of the CASAS dataset and 8.5\% of the MuRAL dataset. To our knowledge, our MDP-based framework is the only Data Association method theoretically capable of mapping a single sensor event to a set of residents. We empirically validate this capability by evaluating the model's precision in cardinality estimation, particularly focusing on its ability to correctly identify the number of residents involved in each event.

Admittedly, achieving high precision in this task remains a formidable challenge. In many instances, the raw binary sensor observations lack sufficient semantic distinctiveness to explicitly reveal the simultaneous presence of multiple residents, making it intrinsically difficult for any model to distinguish a multi-resident activation from a single-resident one. Consequently, as detailed in Table~\ref{tab:bc_multi_resident_event}, the dominant failure mode across all settings is under-assignment, where the model defaults to a single-resident prediction. However, despite these inherent ambiguities, our framework successfully recovers the correct resident cardinality in a significant proportion of cases. These results, while leaving room for improvement, demonstrate the promising potential of our approach to handle complex, concurrent interactions that lie beyond the scope of traditional one-to-one association methods.

Specifically, on the CASAS dataset, the BC model correctly assigns 35.4\% of multi-resident events to the correct set of users. On the MuRAL dataset, both BC and LLM+BC achieve a recognition rate of approximately 41\%. Crucially, the comparison on MuRAL highlights the robustness of the LLM-enhanced approach. While the standard BC model tends to over-assign single-resident events as multi-resident ones (an error rate of 2.8\%), LLM+BC significantly suppresses these false positives, reducing the over-assignment rate to just 0.8\%. This indicates that while both methods possess similar sensitivity in detecting multi-resident activity, LLM+BC offers superior precision, leveraging semantic context to avoid hallucinating multiple residents when only one is present.

\subsection{Cross-Environment Evaluation}
\begin{table}[ht]
\centering
\caption{Cross-environment evaluation of the proposed LLM+BC method on MARBLE~\cite{arrotta2021marble} and MuRAL~\cite{chen2025mural} (train on one dataset and test on the other). Acc and CRL are computed with window size $w=16$.}
\label{tab:llm_bc_cross_env}
\resizebox{0.7\textwidth}{!}{
\begin{tabular}{l|cccc|cccc}
\hline
Dataset     & \multicolumn{4}{c|}{MARBLE~\cite{arrotta2021marble}} & \multicolumn{4}{c}{MuRAL~\cite{chen2025mural}} \\ \hline
Metric      & Acc               & CRL              & BLEU             & RCE              & Acc             & CRL             & BLEU            & RCE            \\ \hline
Without Training       & 0.562             & 3.804   & 0.486            & 1.544   & 0.460  & 2.488           & 0.419           & 1.233          \\
Trained on MuRAL       & 0.765             & 8.853   & 0.760            & 0.727   & 0.773  & 6.034           & 0.799           & 0.261          \\
Trained on MARBLE       & 0.824             & 8.783   & 0.841            & 0.136   & 0.695  & 5.184           & 0.753           & 0.613          \\ \hline
\end{tabular}}
\end{table}

A primary motivation for introducing our LLM+BC framework is to endow the overall method with cross-environment generalization capability. To validate this, we conducted a cross-evaluation experiment across different environments, as detailed in Table~\ref{tab:llm_bc_cross_env}. Specifically, we train LLM+BC on one dataset (source environment), then freeze all model weights and swap the system prompt to the target environment prompt (i.e., the target dataset's resident nomenclature and sensor topology), and finally evaluate the model on the target dataset without any additional fine-tuning. As a control, we also report a no-training ("Without Training") baseline that directly uses Qwen3-4B-Instruct-2507 as the policy in our MDP framework; its relatively low performance (e.g., 0.460 Acc on MuRAL) suggests that directly using a lightweight 4B-parameter LLM as the policy is insufficient for reliable data association, and that domain-specific fine-tuning is necessary for it to adapt to the sensor layouts and resident behaviors of a target smart home.

With this baseline in mind, we next examine cross-environment transfer when the policy is trained in a source environment and evaluated in a target environment. The results reveal a significant asymmetry in transferability: the model trained on the simpler MARBLE dataset generalizes reasonably well to the more complex MuRAL dataset, indicating that the fundamental association logic learned from MARBLE’s high-contrast and mutually exclusive patterns can transfer to more chaotic settings (achieving 0.695 Acc on MuRAL without in-domain training). However, the reverse transfer is much less effective. When trained on the complex MuRAL dataset and evaluated on MARBLE, the model’s behavior degenerates into a naive ``All-to-one'' baseline: despite an Accuracy of 0.765, the exceptionally high RCE (0.727) indicates that most events are indiscriminately assigned to a single resident identity. We hypothesize that this failure is caused by overfitting to the intricate and noisy distributions of MuRAL, which harms the policy's ability to generalize to MARBLE’s simpler, mutually exclusive structure. Further adjusting the number of training epochs on MuRAL may alleviate this issue and lead to different cross-environment transfer results.

\subsection{Impact of Metadata on LLMs}
\begin{table}[t!]
\centering
\caption{Ablation study on the effect of metadata in LLM+BC across three datasets. \textit{LLM+BC w/o Metadata} removes home-specific prompt information and sensor semantics, whereas \textit{LLM+BC w/ Metadata} uses the full prompt. Acc and CRL are calculated with a window size of $w=16$.}
\label{tab:llm_meta}
\resizebox{\textwidth}{!}{
\begin{tabular}{l|cccc|cccc|cccc}
\hline
Dataset       & \multicolumn{4}{c|}{CASAS~\cite{singla2010recognizing}} & \multicolumn{4}{c|}{MARBLE~\cite{arrotta2021marble}} & \multicolumn{4}{c}{MuRAL~\cite{chen2025mural}} \\ \hline
Metric        & Acc                & CRL               & BLEU              & RCE               & Acc               & CRL              & BLEU             & RCE              & Acc             & CRL             & BLEU            & RCE            \\ \hline
BC & 0.757     & 6.234    & 0.727    & 0.021    & 0.792    & 8.710   & 0.797   & 0.136   & 0.728  & 5.476  & 0.774  & 0.437 \\ 
LLM+BC w/o Metadata & 0.761     & 6.135    & 0.725    & 0.024    & 0.789    & 8.037   & 0.786   & 0.182   & 0.756  & 5.942  & 0.777  & 0.632 \\ 
LLM+BC w/ Metadata & /     &  /  & /    & /   & 0.824             & 8.783   & 0.841            & 0.136   & 0.773  & 6.034           & 0.799           & 0.261          \\ \hline
\end{tabular}}
\end{table}

Although the LLM+BC framework demonstrates strong potential, it still relies heavily on environment-specific metadata, and such customized information may not always be available in practical deployments. It is therefore important to examine how the method behaves when these inputs are missing. To this end, we conduct an ablation study in which we remove the House Layout, Sensor Description, and Key Constraints sections from the original system prompt. We further strip home-specific information from the Background and Objective sections, yielding a simplified and unified prompt shared across all datasets. In addition, we anonymize the sensors in each dataset by assigning each sensor an arbitrary numeric identifier with no semantic meaning, and represent every sensor state change uniformly as ON or OFF. Using this prompt and data representation, we retrain the LLM on all three datasets. The results, shown in Table~\ref{tab:llm_meta}, indicate that without metadata, the performance of LLM+BC falls back to a level very close to that of BC. This result is consistent with the design of the ablation, as removing metadata effectively reduces the method from text-enhanced data-driven reasoning to a purely data-driven formulation. Conversely, providing metadata consistently improves performance on MARBLE and MuRAL, confirming that explicit semantic and structural context is a key factor behind the gains of the LLM-enhanced policy.

\subsection{Robustness to Noise}

In real deployments, sensor streams may contain a certain proportion of noisy events caused by sensor faults or other environmental disturbances. Compared with previous methods, our MDP formulation introduces an implicit skip action: when the policy assigns a probability below the multi-label decision threshold (set to 0.5 in our experiments) to every resident slot, the model makes no resident assignment for the current event and simply skips it. We hypothesize that this mechanism provides a natural source of robustness to noise. To test this hypothesis, we inject random noise events into the CASAS dataset at rates of 1\%, 5\%, and 10\%. These noise events are constructed from existing sensors in the environment paired with randomly sampled state changes. During BC training, all resident slots for these noise events are labeled as 0, indicating that the noise belong to no resident. In Table~\ref{tab:robust_noise}, we compare performance degradation under different noise levels against the clean setting, and also report whether the model successfully skips the injected noise. We define \emph{Skip Rate} as the proportion of noise events that are skipped, and \emph{False Skip Rate} as the proportion of valid events that are mistakenly treated as noise and skipped. The results show that, compared with sMRT, our BC method remains highly stable under 1\% noise injection, with almost no performance degradation, while skipping about 10\% of the noise events. When the noise ratio increases to 5\%, the Skip Rate rises to 33\%, while the False Skip Rate remains only 0.7\%, which helps prevent excessive performance degradation. At 10\% noise, however, even skipping roughly half of the noise events is insufficient to prevent noticeable degradation, because a large amount of noise still enters the state history. In addition, the False Skip Rate rises sharply to 11\%, meaning that a substantial portion of valid events is incorrectly discarded, which further harms state construction. Nevertheless, our MDP-based method provides a skip mechanism that is absent from prior approaches, enabling it to maintain stronger performance than previous methods under moderate levels of noise.

\begin{table}[t]
\centering
\caption{Robustness of data association under injected noise on the CASAS dataset. Random noise events are added at different ratios. \emph{Skip} denotes the proportion of injected noise events that are skipped by the model, while \emph{False Skip} denotes the proportion of valid events that are incorrectly skipped.}
\label{tab:robust_noise}
\begin{tabular}{c|c|cccc|cc}
\hline
\multicolumn{1}{l|}{Methods} & Noise\% & Acc                       & CRL                       & BLEU                      & RCE                        & Skip                      & \multicolumn{1}{l}{False Skip} \\ \hline
\multirow{4}{*}{BC}          & 0\%     & 0.757                     & 6.234                     & 0.727                     & 0.021                      & 0.000                     & 0.000                          \\
                             & 1\%     & 0.755                     & 6.201                     & 0.721                     & 0.021                      & 0.103                     & 0.001                          \\
                             & 5\%     & 0.701 & 5.838 & 0.703 & 0.032 & 0.334 & 0.007                          \\
                             & 10\%    & 0.581 & 4.690 & 0.591 & 0.043 & 0.506 & 0.112                          \\ \hline
\multirow{2}{*}{sMRT}        & 0\%     & 0.561                     & 3.051                     & 0.530                     & 1.257                      & /                         & /                              \\
                             & 1\%     & 0.232                     & 1.501                     & 0.190                     & 5.690                      & /                         & /                              \\ \hline
\end{tabular}
\end{table}

\subsection{Simulated Downstream Health Evaluation via Activity Duration Analysis}
\label{sec:casas-duration}

\begin{table}[t]
  \centering\small
  \caption{Per-session activity durations (seconds) for Resident~1 and Resident~2. Each entry is reported as predicted / ground truth. T01--T15 denote the 15 scripted activities defined in the CASAS dataset.}\label{tab:pred-vs-gt}
  
  \textbf{Resident 1}\\[0.5ex]
  \begin{tabular}{@{}lrrrrrrrr@{}}
    \toprule
    Session & T01 & T04 & T06 & T09 & T10 & T12 & T13 & T14 \\
    \midrule
    P19 & \cellcolor{lightpink!14}146 / 170 & 17 / 17 & \cellcolor{lightpink!100}0 / 123 & 411 / 411 & \cellcolor{lightpink!14}92 / 107 & 329 / 329 & \cellcolor{lightpink!7}12 / 13 & \cellcolor{lightpink!1}111 / 112 \\
    P20 & 192 / 192 & 29 / 29 & \cellcolor{lightpink!73}18 / 64 & 348 / 348 & \cellcolor{lightpink!2}84 / 86 & \cellcolor{lightpink!2}427 / 434 & \cellcolor{lightblue!100}226 / 49 & \cellcolor{lightpink!38}59 / 95 \\
    P21 & 247 / 247 & 25 / 25 & \cellcolor{lightpink!47}26 / 48 & 406 / 406 & \cellcolor{lightpink!60}30 / 74 & 211 / 211 & \cellcolor{lightpink!82}5 / 29 & 58 / 58 \\
    P22 & 200 / 200 & \cellcolor{lightpink!2}42 / 43 & \cellcolor{lightpink!48}22 / 41 & 306 / 306 & 65 / 65 & 330 / 330 & \cellcolor{lightblue!100}14 / 2 & \cellcolor{lightpink!54}86 / 187 \\
    P23 & \cellcolor{lightpink!1}301 / 305 & \cellcolor{lightpink!1}31 / 31 & \cellcolor{lightpink!34}84 / 128 & 320 / 320 & 137 / 137 & \cellcolor{lightpink!100}0 / 246 & \cellcolor{lightblue!100}273 / 0 & \cellcolor{lightpink!71}24 / 81 \\
    P24 & 350 / 350 & 36 / 36 & \cellcolor{lightpink!28}72 / 100 & 375 / 375 & 79 / 79 & \cellcolor{lightpink!27}406 / 554 & \cellcolor{lightpink!99}1 / 88 & \cellcolor{lightpink!4}94 / 97 \\
    P25 & \cellcolor{lightpink!18}311 / 379 & 65 / 65 & \cellcolor{lightpink!27}112 / 153 & \cellcolor{lightpink!83}67 / 392 & \cellcolor{lightpink!100}0 / 81 & \cellcolor{lightpink!73}138 / 505 & \cellcolor{lightblue!87}290 / 155 & \cellcolor{lightpink!28}77 / 106 \\
    P26 & 312 / 312 & 50 / 50 & \cellcolor{lightpink!37}54 / 85 & 347 / 347 & \cellcolor{lightpink!85}13 / 89 & 436 / 436 & \cellcolor{lightblue!100}84 / 16 & \cellcolor{lightpink!8}86 / 93 \\
    \midrule
    Our Err \% & -4.5\% & -0.3\% & -47.7\% & -11.2\% & -30.4\% & -25.2\% & +157.1\% & -28.2\% \\
    sMRT Err \% & -72.7\% & -45.4\% & -78.9\% & -55.3\% & -87.0\% & -53.7\% & +12.2\% & -98.7\% \\
    \bottomrule
  \end{tabular}

  \vspace{0.5em} %

  \textbf{Resident 2}\\[0.5ex]
  \begin{tabular}{@{}lrrrrrrrr@{}}
    \toprule
    Session & T02 & T03 & T05 & T07 & T08 & T11 & T13 & T15 \\
    \midrule
    P19 & \cellcolor{lightpink!12}64 / 73 & 20 / 20 & \cellcolor{lightpink!5}67 / 70 & \cellcolor{lightpink!68}17 / 55 & \cellcolor{lightpink!100}0 / 340 & \cellcolor{lightpink!14}119 / 140 & \cellcolor{lightpink!26}244 / 329 & \cellcolor{lightpink!2}113 / 116 \\
    P20 & 83 / 83 & 18 / 18 & 26 / 26 & \cellcolor{lightpink!55}62 / 139 & 267 / 267 & \cellcolor{lightpink!2}96 / 98 & 351 / 351 & \cellcolor{lightpink!32}67 / 98 \\
    P21 & \cellcolor{lightpink!5}51 / 53 & 17 / 17 & 86 / 86 & \cellcolor{lightpink!32}28 / 41 & 345 / 345 & \cellcolor{lightpink!5}125 / 131 & \cellcolor{lightpink!21}173 / 218 & \cellcolor{lightpink!20}55 / 68 \\
    P22 & 103 / 103 & \cellcolor{lightpink!14}36 / 42 & \cellcolor{lightpink!59}20 / 48 & \cellcolor{lightpink!7}48 / 52 & \cellcolor{lightpink!36}140 / 217 & 66 / 66 & \cellcolor{lightpink!29}238 / 337 & \cellcolor{lightpink!38}115 / 184 \\
    P23 & 70 / 70 & 24 / 24 & \cellcolor{lightpink!4}179 / 186 & \cellcolor{lightpink!8}112 / 122 & \cellcolor{lightpink!3}306 / 314 & \cellcolor{lightpink!91}11 / 127 & \cellcolor{lightpink!100}0 / 273 & \cellcolor{lightpink!56}37 / 84 \\
    P24 & 92 / 92 & 45 / 45 & \cellcolor{lightpink!2}233 / 238 & \cellcolor{lightpink!28}104 / 143 & \cellcolor{lightpink!100}0 / 328 & \cellcolor{lightpink!4}88 / 91 & \cellcolor{lightpink!29}426 / 600 & \cellcolor{lightpink!74}29 / 111 \\
    P25 & \cellcolor{lightpink!8}88 / 96 & 61 / 61 & 134 / 134 & \cellcolor{lightpink!5}211 / 221 & \cellcolor{lightpink!11}289 / 325 & \cellcolor{lightpink!88}14 / 115 & \cellcolor{lightpink!68}150 / 463 & \cellcolor{lightpink!55}49 / 109 \\
    P26 & \cellcolor{lightpink!3}191 / 197 & \cellcolor{lightpink!29}34 / 47 & 38 / 38 & \cellcolor{lightpink!7}165 / 177 & 283 / 283 & \cellcolor{lightpink!83}16 / 94 & \cellcolor{lightpink!35}300 / 459 & \cellcolor{lightpink!14}81 / 94 \\
    \midrule
    Our Err \% & -3.3\% & -6.9\% & -5.2\% & -21.4\% & -32.6\% & -37.9\% & -37.9\% & -36.8\% \\
    sMRT Err \% & -55.7\% & -42.2\% & -60.7\% & -83.8\% & -45.2\% & -82.5\% & -71.4\% & -94.5\% \\
    \bottomrule
  \end{tabular}
\end{table}

Finally, we return to the broader motivation of this work: to extend smart home applications that were previously developed for single-resident ambient sensor environments into multi-resident settings in an algorithmic manner. To this end, we finally verify the practical potential of our framework for downstream health-related applications. Following Aramendi et al.~\cite{aramendi2018automatic}, who assessed residents' health status from activity durations in single-resident CASAS environments, we ask whether a similar pipeline can be extended to the corresponding two-resident CASAS setting using the sequences separated by our method. To isolate the effect of data association, we assume 100\% HAR accuracy on the separated single-resident sequences, i.e., we directly use the ground-truth activity labels. This simplification avoids introducing additional noise from HAR, which lies outside the main scope of our contribution, and is also reasonable because Aramendi et al. reported over 98\% HAR accuracy on single-resident CASAS sequences. We then compare predicted and ground-truth activity durations at the $(\textrm{resident},\textrm{activity})$ level. On the ground-truth side, each activity duration is computed directly from the CASAS annotations. On the prediction side, we align the session-local resident labels produced by our method with the CASAS resident IDs using the first event of the session, and then apply this mapping to all subsequent events to reconstruct the predicted $(r,a)$ segments. For each resident--activity pair, we finally compare the predicted duration against the ground truth.

Table~\ref{tab:pred-vs-gt} reports the per-session activity durations for each resident on the test sessions P19--P26, where T01--T15 denote the 15 scripted ADL tasks in the CASAS protocol. Overall, compared with the fragmented trajectories produced by sMRT, our method substantially reduces activity-duration estimation errors, with the clearest improvements appearing in T01--T05, which occur in the early part of every session before substantial error accumulation sets in. This improvement further suggests the practical potential of our method for supporting duration-based downstream health analysis.

However, the more noticeable remaining discrepancies tend to appear in later activities (T06--T15), where two factors compound the difficulty: first, error accumulation over time makes identity swaps more likely; second, these activities often involve close interactions and overlapping sensor activations between the two residents, which further increases the difficulty of distinguishing them. In particular, Resident~1's T06 and Resident~2's T07 are difficult to distinguish, and a similar pattern appears for Resident~2's T08 and Resident~1's T09 during the dinner-preparation stage. These activity pairs often occur in parallel, exhibit substantial spatial overlap, and involve many multi-resident sensor events. For example, in P19 and P24, Resident~2's T08 is predicted entirely as Resident~1's T09, indicating that the two residents' trajectories have effectively been merged into a single track; this again reflects the difficulty of handling multi-resident events. A related source of error appears in T13, which is primarily carried out by Resident~2 but includes brief assistance from Resident~1. Because this activity contains a short but meaningful overlap between the two residents, the model may either absorb Resident~1's contribution into Resident~2's main trajectory or over-extend Resident~1's involvement, thereby over-predicting Resident~1's T13 duration. P23 provides a representative example of a longer-horizon failure mode. In this session, Resident~1's T12 and T13 become fully entangled with Resident~2's T11 and T13. At the onset of T11 and T12, the two residents are assigned in a swapped manner, and this identity reversal then persists through T13 and continues until T14 and T15. Thus, the dominant error is not an isolated duration mismatch within a single activity, but a sustained identity swap over the later part of the session that systematically exchanges the two residents' trajectories. Taken together, these results suggest that the remaining duration errors are driven less by isolated activity recognition failures than by ambiguous multi-resident interactions and long-horizon identity swaps in the later parts of the session.

\section{Limitations and Future Work}
Our work establishes a technical foundation for extending smart-home ambient sensor applications from single-resident settings to multi-resident environments by formulating Data Association as an MDP and solving it through BC with both Transformer-based and LLM-based policies. Across all three datasets, the proposed framework consistently outperforms prior baselines in tracking consistency, resident-count estimation, overlapping-event attribution, robustness to noise, and cross-environment generalizability. These improvements appear to arise from two main factors: first, the MDP formulation captures global spatiotemporal consistency over assignment histories; second, the BC paradigm enables data-driven and effective policy learning without requiring carefully designed reward functions. Consequently, the framework mitigates the severe trajectory fragmentation often observed in filtering-based methods such as sMRT, while also allowing deferred decision-making, in which later observations help resolve earlier ambiguities during inference. Moreover, experiments with LLM-based policies indicate that the proposed training paradigm not only surpasses prior prompt-based methods such as LAHAR, but also retains useful association capability under reduced metadata conditions. Finally, the simulated downstream health-analysis experiment suggests that improved DA can substantially reduce activity-duration estimation errors in multi-resident settings, highlighting the broader potential of this framework as an enabling layer for future smart-home applications. Nevertheless, the results also reveal several important limitations that must be addressed before such methods can be robustly deployed at scale.

\subsection{Error Accumulation and Distribution Shift}
Although our supervised Behavioral Cloning (BC) approach is effective, our experimental analysis reveals a susceptibility to error accumulation over long horizons. As demonstrated by the window-size analysis in Section~\ref{sec:error_accumulation}, tracking performance degrades notably as the evaluation window increases. This indicates that early identity assignment mistakes are not isolated events; rather, they propagate through the maintained resident histories, causing the agent to drift further from the correct trajectory. Theoretically, this phenomenon stems from the \textit{covariate shift} inherent in sequential decision-making. Standard BC relies on the assumption of independent and identically distributed (i.i.d.) data, training the policy exclusively on perfect expert trajectories. Consequently, when the agent encounters ``corrupted'' states caused by its own prediction errors during inference, it lacks the capability to recover. To mitigate this issue, we encourage future research to explore two complementary directions: \textit{training-side robustness} and \textit{deployment-side lifecycle management}.

From the training perspective, adopting Dataset Aggregation (DAgger)~\cite{ross2011reduction} offers a promising solution. Unlike static BC, DAgger allows the learner policy to execute actions during training, naturally generating deviant system states. By querying the expert for correct assignments given these corrupted histories, the policy explicitly learns to recover from tracking errors. This interactive process effectively trains the agent to re-associate events correctly even after misallocations, thereby enhancing the policy's intrinsic robustness against distribution shift.

Beyond imitation-learning extensions, an important next step is to build on the strong BC baseline with long-horizon Reinforcement Learning. The BC-trained policy provides a stable warm start, after which the agent can sample its own trajectories and be optimized for trajectory-level association quality. This direction is also supported by our findings in Section~\ref{sec: exp_beam_search}: the gains from trajectory-level search suggest that preserving multiple candidate futures can already mitigate error accumulation at inference time. RL offers a natural way to move this benefit into training, shifting the objective from matching expert actions locally to improving the global consistency of the full attribution sequence. Rather than relying on dense heuristic rewards, a more principled design could use delayed rewards over blocks of events together with a value model that estimates the utility of intermediate states. A further extension is to enlarge the action space itself so that the policy is not restricted to irreversible event-to-resident assignments. Besides assigning the current event, the agent could invoke temporary \textit{skip} actions or history-editing actions that revise earlier attributions when later evidence reveals an inconsistency. This idea is especially natural for the LLM-based policy, which could emit structured edit commands over the resident history. Overall, BC initialization, beam-search inference, and subsequent RL fine-tuning---potentially with an expanded action space for corrective decisions---form a natural path toward more stable long-horizon reasoning.

From the deployment perspective, future work may benefit from system-level lifecycle management inspired by classical Multi-Target Tracking (MTT) theory~\cite{blackman1999design}. Maintaining resident states indefinitely can risk permanent divergence, similar to the kidnapped robot problem~\cite{thrun2005probabilistic}. We believe an Error-Triggered Re-initialization mechanism may be useful: by monitoring uncertainty metrics, the system can detect out-of-distribution states and perform a hard reset to truncate error propagation. To reduce track fragmentation, one practical strategy may be to execute such a reset when the environment is inferred to be vacant, for example, when the house is inferred to be empty.

\subsection{Multi-Resident Events}
An important advantage of our proposed method is applying multi-label classifier, which, to our knowledge, is the first Data Association model that can assign a single ambient sensor event to multiple residents. However, as shown in Section~\ref{sec:multi_events}, although our method can correctly recover the resident cardinality for a subset of multi-resident events, a substantial fraction is still under-assigned and treated as single-resident events.

We believe this limitation is primarily driven by the data rather than the model formulation. Truly concurrent events are both intrinsically ambiguous under binary ambient sensors and relatively rare in the available datasets (accounting for approximately 10\% in MuRAL and CASAS), which limits the evidence available for learning and evaluation. Future work could therefore prioritize data augmentation via trajectory mixing to synthesize additional multi-resident overlaps and strengthen the model's sensitivity to concurrent activations.

\subsection{Datasets Limitations}

Although our work has been comprehensively evaluated across three datasets with diverse characteristics, we must acknowledge the inherent domain gap between these controlled experimental setups and in-the-wild smart home environments. The primary discrepancy lies in sensor topology and deployment density. As introduced in Section~\ref{sec: datasets}, the CASAS environment features a high-density array of motion sensors mounted on the ceiling. Such a high-cost and technically demanding deployment differs drastically from the reality of consumer smart homes. Similarly, while MARBLE and MuRAL more closely resemble modern smart home configurations, they still utilize over 16 sensors and employ highly localized, custom deployment strategies designed specifically to isolate individuals (e.g., the seating area in MARBLE is densely equipped with eight distinct pressure mats to ensure the exclusive capture of a single occupant's movements). Such fine-grained sensor layouts are exceedingly rare in typical households. Secondly, our method faces a steep barrier regarding annotation costs. Under the current supervised learning paradigm, our proposed approach relies heavily on event-level identity annotations. Acquiring such detailed labels in everyday home environments without professional volunteers is prohibitively expensive and highly impractical. Consequently, the actual generalization capability and deployment performance of our method in natural, resource-constrained households with sparse sensor configurations remain to be further validated in future work.

\subsection{Cross-Environment Generalization and Reasoning}
Although our LLM+BC framework exhibits promising cross-dataset transfer, relying on direct input-to-output mapping limits its robustness against severe domain shifts. Without explicit reasoning, the policy risks overfitting to superficial statistical correlations specific to the training sensors, rather than learning the invariant causal logic required for data association. We hypothesize that incorporating \textit{Chain-of-Thought (CoT)}~\cite{wei2022chain} is crucial for bridging this gap. To elicit the latent reasoning capabilities of LLMs and achieve true generalizability, future work should explore \textit{Reasoning-Aware Fine-tuning} through Reasoning Distillation via Rejection Sampling~\cite{zelikman2022star} and Outcome-Supervised Reinforcement Learning~\cite{guo2025deepseek}.

\subsection{Scalability and Real-time System Integration}
Beyond the domain gaps discussed above, translating our theoretical framework into an industrial smart home foundation requires addressing additional practical considerations: scalability concerning the number of residents and integration into real-time streaming systems.

Regarding scalability, as the number of household members increases, the ambiguity of data association and the corresponding search space typically experience exponential growth. However, our MDP formulation is logically scalable: introducing new residents merely results in a linear expansion of the action space, which our Transformer/LLM-based policy can naturally accommodate through contextual sequence modeling. Computationally, beam search effectively serves to cap the processing overhead. By strictly bounding the number of maintained hypotheses to a constant beam width $B$, the system bypasses the combinatorial explosion of exhaustive search, ensuring that the inference complexity per step scales linearly rather than exponentially with the number of residents.

Regarding system integration, our method is designed to function as a low-level data processing middleware seamlessly plugged into real-time streaming smart home ecosystems. In practice, as asynchronous sensor events arrive, the system maintains a sliding temporal window and a state buffer. For unambiguous events, the module yields immediate assignments to ensure low latency. When facing highly ambiguous scenarios, it triggers a ``deferred decision'', leveraging the buffer to cross-validate and prune hypotheses as new streaming data arrives. In practice, many downstream applications, such as health monitoring and behavioral analysis, do not require extremely strict latency and therefore align well with our framework, which can trade a small amount of delay for better long-horizon consistency. For applications that do require near-real-time responses, such as home security anomaly detection or smart-home automation, the Transformer-based policy can fall back to greedy online inference to ensure low latency. When an LLM is used as the policy, deployment is also increasingly practical: models at the 4B scale can already be run locally on commodity devices such as a Mac mini with sufficiently fast inference for system integration, allowing the policy to be exposed as an API within a smart-home pipeline.

\section{Conclusions}
\label{sect:conclusion}
In this paper, we presented a framework to address the challenge of Data Association in multi-resident smart home environments. 
We identified that existing approaches, whether relying on event-centric classification or Bayesian filtering, are fundamentally limited by the rigid first-order Markov assumption and the lack of a persistent global memory. 
These limitations frequently result in inconsistent assignments for silent residents and trajectory fragmentation in complex scenarios.

To overcome these structural bottlenecks, we reformulated the Data Association problem as a sequential decision-making task within a Markov Decision Process (MDP). 
This formulation allows us to explicitly represent the global assignment history and reason over longer behavioral context, although our experiments also show that robust long-horizon tracking remains challenging because early mistakes can accumulate under Behavior Cloning.
By solving this MDP through Behavior Cloning, we adopt a fully data-driven approach that avoids handcrafted reward design and the tractability constraints of traditional particle filters, but currently still depends on environment-specific supervision for reliable adaptation.
Our framework was instantiated with two complementary policy architectures: a lightweight Transformer that handles association as a multi-label classification task, and a fine-tuned Large Language Model that reformulates the problem as conditional text generation to leverage pre-trained semantic knowledge.

Extensive evaluations on the CASAS, MARBLE, and MuRAL datasets demonstrate that our proposed method significantly outperforms state-of-the-art baselines on these benchmarks.
Our results specifically highlight that the framework improves global consistency relative to prior methods and is particularly effective at resolving concurrent events involving multiple residents.
Furthermore, the LLM-based policy shows encouraging cross-environment transfer under controlled benchmark settings, suggesting that leveraging semantic reasoning is a promising direction, while its robustness under sparse, outdated, or changing metadata and under real household deployment conditions remains unresolved.

Overall, we view this work as a technical foundation rather than a deployment-ready solution. Moving toward real-world smart-home deployment will require addressing long-horizon error accumulation, robustness to delayed and missing sensor data, dependence on static metadata, annotation cost, and validation in more natural household environments.

\bibliographystyle{plainnat}
\bibliography{sample-base}

\begin{thebibliography}{45}
\providecommand{\natexlab}[1]{#1}
\providecommand{\url}[1]{\texttt{#1}}
\expandafter\ifx\csname urlstyle\endcsname\relax
  \providecommand{\doi}[1]{doi: #1}\else
  \providecommand{\doi}{doi: \begingroup \urlstyle{rm}\Url}\fi

\bibitem[Acampora et~al.(2013)Acampora, Cook, Rashidi, and
  Vasilakos]{acampora2013survey}
Giovanni Acampora, Diane~J Cook, Parisa Rashidi, and Athanasios~V Vasilakos.
\newblock A survey on ambient intelligence in healthcare.
\newblock \emph{Proceedings of the IEEE}, 101\penalty0 (12):\penalty0
  2470--2494, 2013.

\bibitem[Afroz et~al.(2025)Afroz, Nyakwende, and Goswami]{afroz2025intrusion}
Md~Afroz, Emmanuel Nyakwende, and Birendra Goswami.
\newblock Intrusion detection in smart home environments: A machine learning
  approach.
\newblock \emph{Transportation Research Procedia}, 84:\penalty0 243--250, 2025.

\bibitem[Ahmadi-Karvigh et~al.(2018)Ahmadi-Karvigh, Ghahramani, Becerik-Gerber,
  and Soibelman]{ahmadi2018real}
Simin Ahmadi-Karvigh, Ali Ghahramani, Burcin Becerik-Gerber, and Lucio
  Soibelman.
\newblock Real-time activity recognition for energy efficiency in buildings.
\newblock \emph{Applied energy}, 211:\penalty0 146--160, 2018.

\bibitem[Alemdar et~al.(2013)Alemdar, Ertan, Incel, and Ersoy]{alemdar2013aras}
Hande Alemdar, Halil Ertan, Ozlem~Durmaz Incel, and Cem Ersoy.
\newblock Aras human activity datasets in multiple homes with multiple
  residents.
\newblock In \emph{2013 7th International Conference on Pervasive Computing
  Technologies for Healthcare and Workshops}, pages 232--235. IEEE, 2013.

\bibitem[Aramendi et~al.(2018)Aramendi, Weakley, Goenaga, Schmitter-Edgecombe,
  and Cook]{aramendi2018automatic}
Ane~Alberdi Aramendi, Alyssa Weakley, Asier~Aztiria Goenaga, Maureen
  Schmitter-Edgecombe, and Diane~J Cook.
\newblock Automatic assessment of functional health decline in older adults
  based on smart home data.
\newblock \emph{Journal of biomedical informatics}, 81:\penalty0 119--130,
  2018.

\bibitem[Arrotta et~al.(2021)Arrotta, Bettini, and
  Civitarese]{arrotta2021marble}
Luca Arrotta, Claudio Bettini, and Gabriele Civitarese.
\newblock The marble dataset: Multi-inhabitant activities of daily living
  combining wearable and environmental sensors data.
\newblock In \emph{International Conference on Mobile and Ubiquitous Systems:
  Computing, Networking, and Services}, pages 451--468, Cham, 2021. Springer,
  Springer International Publishing.

\bibitem[Arrotta et~al.(2024)Arrotta, Civitarese, Chen, Cumin, and
  Bettini]{arrotta2024multi}
Luca Arrotta, Gabriele Civitarese, Xi~Chen, Julien Cumin, and Claudio Bettini.
\newblock Multi-subject human activities: A survey of recognition and
  evaluation methods based on a formal framework.
\newblock \emph{Expert Systems with Applications}, page 126178, 2024.

\bibitem[Benmansour et~al.(2015)Benmansour, Bouchachia, and
  Feham]{benmansour2015multioccupant}
Asma Benmansour, Abdelhamid Bouchachia, and Mohammed Feham.
\newblock Multioccupant activity recognition in pervasive smart home
  environments.
\newblock \emph{ACM Computing Surveys (CSUR)}, 48\penalty0 (3):\penalty0 1--36,
  2015.

\bibitem[Blackman and Popoli(1999)]{blackman1999design}
Samuel~S Blackman and Robert Popoli.
\newblock \emph{Design and analysis of modern tracking systems}.
\newblock Artech House, 1999.

\bibitem[Chan et~al.(2024)Chan, Cai, Qian, Coutts, Phan, Gregson, Lipsett,
  Rinc{\'o}n, et~al.]{chan2024home}
Andrew Chan, Joanne Cai, Linna Qian, Brendan Coutts, Steven Phan, Geoff
  Gregson, Michael Lipsett, Adriana M~R{\'\i}os Rinc{\'o}n, et~al.
\newblock In-home positioning for remote home health monitoring in older
  adults: Systematic review.
\newblock \emph{JMIR aging}, 7\penalty0 (1):\penalty0 e57320, 2024.

\bibitem[Chen and Tong(2014)]{chen2014two}
Rong Chen and Yu~Tong.
\newblock A two-stage method for solving multi-resident activity recognition in
  smart environments.
\newblock \emph{Entropy}, 16\penalty0 (4):\penalty0 2184--2203, 2014.

\bibitem[Chen et~al.(2024{\natexlab{a}})Chen, Cumin, Ramparany, and
  Vaufreydaz]{chen2024generative}
Xi~Chen, Julien Cumin, Fano Ramparany, and Dominique Vaufreydaz.
\newblock Generative resident separation and multi-label classification for
  multi-person activity recognition.
\newblock In \emph{2024 IEEE International Conference on Pervasive Computing
  and Communications Workshops and other Affiliated Events (PerCom Workshops)},
  pages 1--6, Biarritz, France, 2024{\natexlab{a}}. IEEE, IEEE.

\bibitem[Chen et~al.(2024{\natexlab{b}})Chen, Cumin, Ramparany, and
  Vaufreydaz]{chen2024towards}
Xi~Chen, Julien Cumin, Fano Ramparany, and Dominique Vaufreydaz.
\newblock Towards llm-powered ambient sensor based multi-person human activity
  recognition.
\newblock In \emph{2024 IEEE 30th International Conference on Parallel and
  Distributed Systems (ICPADS)}, pages 609--616, Belgrade, Serbia,
  2024{\natexlab{b}}. IEEE, IEEE.

\bibitem[Chen et~al.(2025)Chen, Cumin, Ramparany, and
  Vaufreydaz]{chen2025mural}
Xi~Chen, Julien Cumin, Fano Ramparany, and Dominique Vaufreydaz.
\newblock {MuRAL: A Multi-Resident Ambient Sensor Dataset Annotated with
  Natural Language for Activities of Daily Living}.
\newblock \emph{HAL preprint hal-05048859}, 2025.
\newblock URL \url{https://hal.science/hal-05048859}.

\bibitem[Chiang et~al.(2010)Chiang, Hsu, Lu, Fu, and
  Hsu]{chiang2010interaction}
Yi-Ting Chiang, Kuo-Chung Hsu, Ching-Hu Lu, Li-Chen Fu, and Jane Yung-Jen Hsu.
\newblock Interaction models for multiple-resident activity recognition in a
  smart home.
\newblock In \emph{2010 IEEE/RSJ International Conference on Intelligent Robots
  and Systems}, pages 3753--3758. IEEE, 2010.

\bibitem[Cook et~al.(2010)Cook, Crandall, Singla, and
  Thomas]{cook2010detection}
Diane~J Cook, Aaron Crandall, Geetika Singla, and Brian Thomas.
\newblock Detection of social interaction in smart spaces.
\newblock \emph{Cybernetics and Systems: An International Journal}, 41\penalty0
  (2):\penalty0 90--104, 2010.

\bibitem[Crandall and Cook(2008{\natexlab{a}})]{crandall2008attributing}
Aaron~S Crandall and Diane Cook.
\newblock Attributing events to individuals in multi-inhabitant environments.
\newblock In \emph{2008 IET 4th International Conference on Intelligent
  Environments}, pages 1--8. IET, 2008{\natexlab{a}}.

\bibitem[Crandall and Cook(2008{\natexlab{b}})]{crandall2008resident}
Aaron~S Crandall and Diane~J Cook.
\newblock Resident and caregiver: Handling multiple people in a smart care
  facility.
\newblock In \emph{AAAI Fall symposium: AI in eldercare: new solutions to old
  problems}, pages 39--47, 2008{\natexlab{b}}.

\bibitem[Crandall and Cook(2010)]{crandall2010using}
Aaron~S Crandall and Diane~J Cook.
\newblock Using a hidden markov model for resident identification.
\newblock In \emph{2010 Sixth International Conference on Intelligent
  Environments}, pages 74--79. IEEE, 2010.

\bibitem[F{\"a}hrmann et~al.(2024)F{\"a}hrmann, Mart{\'\i}n, S{\'a}nchez, and
  Damer]{fahrmann2024anomaly}
Daniel F{\"a}hrmann, Laura Mart{\'\i}n, Luis S{\'a}nchez, and Naser Damer.
\newblock Anomaly detection in smart environments: a comprehensive survey.
\newblock \emph{IEEE access}, 2024.

\bibitem[Giannios et~al.(2024)Giannios, Mpaltadoros, Alepopoulos,
  Grammatikopoulou, Stavropoulos, Nikolopoulos, Lazarou, Tsolaki, and
  Kompatsiaris]{giannios2024semantic}
Giorgos Giannios, Lampros Mpaltadoros, Vasilis Alepopoulos, Margarita
  Grammatikopoulou, Thanos~G Stavropoulos, Spiros Nikolopoulos, Ioulietta
  Lazarou, Magda Tsolaki, and Ioannis Kompatsiaris.
\newblock A semantic framework to detect problems in activities of daily living
  monitored through smart home sensors.
\newblock \emph{Sensors}, 24\penalty0 (4):\penalty0 1107, 2024.

\bibitem[Guo et~al.(2025)Guo, Yang, Zhang, Song, Wang, Zhu, Xu, Zhang, Ma, Bi,
  et~al.]{guo2025deepseek}
Daya Guo, Dejian Yang, Haowei Zhang, Junxiao Song, Peiyi Wang, Qihao Zhu,
  Runxin Xu, Ruoyu Zhang, Shirong Ma, Xiao Bi, et~al.
\newblock Deepseek-r1 incentivizes reasoning in llms through reinforcement
  learning.
\newblock \emph{Nature}, 645\penalty0 (8081):\penalty0 633--638, 2025.

\bibitem[Hayward et~al.(2022)Hayward, van Lopik, Hinde, and
  West]{hayward2022survey}
SJ~Hayward, Katherine van Lopik, Christopher Hinde, and Andrew~A West.
\newblock A survey of indoor location technologies, techniques and applications
  in industry.
\newblock \emph{Internet of Things}, 20:\penalty0 100608, 2022.

\bibitem[Hsu et~al.(2010)Hsu, Chiang, Lin, Lu, Hsu, and Fu]{hsu2010strategies}
Kuo-Chung Hsu, Yi-Ting Chiang, Gu-Yang Lin, Ching-Hu Lu, Jane Yung-Jen Hsu, and
  Li-Chen Fu.
\newblock Strategies for inference mechanism of conditional random fields for
  multiple-resident activity recognition in a smart home.
\newblock In \emph{Trends in Applied Intelligent Systems: 23rd International
  Conference on Industrial Engineering and Other Applications of Applied
  Intelligent Systems, IEA/AIE 2010, Cordoba, Spain, June 1-4, 2010,
  Proceedings, Part I 23}, pages 417--426. Springer, 2010.

\bibitem[Hu et~al.(2022)Hu, Shen, Wallis, Allen-Zhu, Li, Wang, Wang, Chen,
  et~al.]{hu2022lora}
Edward~J Hu, Yelong Shen, Phillip Wallis, Zeyuan Allen-Zhu, Yuanzhi Li, Shean
  Wang, Lu~Wang, Weizhu Chen, et~al.
\newblock Lora: Low-rank adaptation of large language models.
\newblock \emph{ICLR}, 1\penalty0 (2):\penalty0 3, 2022.

\bibitem[Lashkari et~al.(2019)Lashkari, Chen, and Musilek]{lashkari2019energy}
Behzad Lashkari, Yuxiang Chen, and Petr Musilek.
\newblock Energy management for smart homes—state of the art.
\newblock \emph{Applied Sciences}, 9\penalty0 (17):\penalty0 3459, 2019.

\bibitem[Lin(2024)]{lin2024behavioral}
Baihan Lin.
\newblock Behavioral cloning and imitation learning.
\newblock In \emph{Reinforcement Learning Methods in Speech and Language
  Technology}, pages 63--67. Springer, 2024.

\bibitem[Mahmud et~al.(2024)Mahmud, Hussain, and Karamat]{mahmud2024towards}
Umar Mahmud, Shariq Hussain, and Tehmina Karamat.
\newblock Towards an intelligent energy conservation approach for context-aware
  systems in smart environments.
\newblock \emph{Frontiers in Computer Science}, 6:\penalty0 1525382, 2024.

\bibitem[Papineni et~al.(2002)Papineni, Roukos, Ward, and
  Zhu]{papineni2002bleu}
Kishore Papineni, Salim Roukos, Todd Ward, and Wei-Jing Zhu.
\newblock Bleu: a method for automatic evaluation of machine translation.
\newblock In \emph{Proceedings of the 40th annual meeting of the Association
  for Computational Linguistics}, pages 311--318, USA, 2002. Association for
  Computational Linguistics.

\bibitem[Puterman(2014)]{puterman2014markov}
Martin~L Puterman.
\newblock \emph{Markov decision processes: discrete stochastic dynamic
  programming}.
\newblock John Wiley \& Sons, 2014.

\bibitem[Rahim et~al.(2023)Rahim, Zhong, Ahmad, Ahmad, P{\l}awiak, and
  Hammad]{rahim2023enhancing}
Asif Rahim, Yanru Zhong, Tariq Ahmad, Sadique Ahmad, Pawe{\l} P{\l}awiak, and
  Mohamed Hammad.
\newblock Enhancing smart home security: anomaly detection and face recognition
  in smart home iot devices using logit-boosted cnn models.
\newblock \emph{Sensors}, 23\penalty0 (15):\penalty0 6979, 2023.

\bibitem[Rahman et~al.(2024)Rahman, Gupta, Bhatt, Shokouhmand, and
  Faezipour]{rahman2024comprehensive}
Md~Motiur Rahman, Deepti Gupta, Smriti Bhatt, Shiva Shokouhmand, and Miad
  Faezipour.
\newblock A comprehensive review of machine learning approaches for anomaly
  detection in smart homes: experimental analysis and future directions.
\newblock \emph{Future Internet}, 16\penalty0 (4):\penalty0 139, 2024.

\bibitem[Riboni and Murru(2020)]{riboni2020unsupervised}
Daniele Riboni and Flavia Murru.
\newblock Unsupervised recognition of multi-resident activities in smart-homes.
\newblock \emph{IEEE Access}, 8:\penalty0 201985--201994, 2020.

\bibitem[Riboni et~al.(2016)Riboni, Bettini, Civitarese, Janjua, and
  Helaoui]{riboni2016smartfaber}
Daniele Riboni, Claudio Bettini, Gabriele Civitarese, Zaffar~Haider Janjua, and
  Rim Helaoui.
\newblock Smartfaber: Recognizing fine-grained abnormal behaviors for early
  detection of mild cognitive impairment.
\newblock \emph{Artificial intelligence in medicine}, 67:\penalty0 57--74,
  2016.

\bibitem[Ross et~al.(2011)Ross, Gordon, and Bagnell]{ross2011reduction}
St{\'e}phane Ross, Geoffrey Gordon, and Drew Bagnell.
\newblock A reduction of imitation learning and structured prediction to
  no-regret online learning.
\newblock In \emph{Proceedings of the fourteenth international conference on
  artificial intelligence and statistics}, pages 627--635. JMLR Workshop and
  Conference Proceedings, 2011.

\bibitem[Singla et~al.(2010)Singla, Cook, and
  Schmitter-Edgecombe]{singla2010recognizing}
Geetika Singla, Diane~J Cook, and Maureen Schmitter-Edgecombe.
\newblock Recognizing independent and joint activities among multiple residents
  in smart environments.
\newblock \emph{Journal of ambient intelligence and humanized computing},
  1:\penalty0 57--63, 2010.

\bibitem[Thrun et~al.(2005)Thrun, Burgard, and Fox]{thrun2005probabilistic}
Sebastian Thrun, Wolfram Burgard, and Dieter Fox.
\newblock \emph{Probabilistic robotics}.
\newblock MIT press, 2005.

\bibitem[Vaswani et~al.(2017)Vaswani, Shazeer, Parmar, Uszkoreit, Jones, Gomez,
  Kaiser, and Polosukhin]{vaswani2017attention}
Ashish Vaswani, Noam Shazeer, Niki Parmar, Jakob Uszkoreit, Llion Jones,
  Aidan~N Gomez, {\L}ukasz Kaiser, and Illia Polosukhin.
\newblock Attention is all you need.
\newblock \emph{Advances in neural information processing systems}, 30, 2017.

\bibitem[Wang and Cook(2020)]{wang2020smrt}
Tinghui Wang and Diane~J Cook.
\newblock smrt: Multi-resident tracking in smart homes with sensor
  vectorization.
\newblock \emph{IEEE transactions on pattern analysis and machine
  intelligence}, 43\penalty0 (8):\penalty0 2809--2821, 2020.

\bibitem[Wang and Cook(2021)]{wang2021multi}
Tinghui Wang and Diane~J Cook.
\newblock Multi-person activity recognition in continuously monitored smart
  homes.
\newblock \emph{IEEE transactions on emerging topics in computing}, 10\penalty0
  (2):\penalty0 1130--1141, 2021.

\bibitem[Wei et~al.(2022)Wei, Wang, Schuurmans, Bosma, Xia, Chi, Le, Zhou,
  et~al.]{wei2022chain}
Jason Wei, Xuezhi Wang, Dale Schuurmans, Maarten Bosma, Fei Xia, Ed~Chi, Quoc~V
  Le, Denny Zhou, et~al.
\newblock Chain-of-thought prompting elicits reasoning in large language
  models.
\newblock \emph{Advances in neural information processing systems},
  35:\penalty0 24824--24837, 2022.

\bibitem[Wilson and Atkeson(2005)]{wilson2005simultaneous}
Daniel~H Wilson and Chris Atkeson.
\newblock Simultaneous tracking and activity recognition (star) using many
  anonymous, binary sensors.
\newblock In \emph{International conference on pervasive computing}, pages
  62--79. Springer, 2005.

\bibitem[Wu et~al.(2024)Wu, Chen, Lin, Wen, and Chen]{wu2024smart}
Zhouwen Wu, Xia Chen, Yujun Lin, Jinyu Wen, and Yin Chen.
\newblock A smart home energy management system based on human activity
  recognition and deep reinforcement learning.
\newblock \emph{Energy and Buildings}, 325:\penalty0 114951, 2024.

\bibitem[Zelikman et~al.(2022)Zelikman, Wu, Mu, and Goodman]{zelikman2022star}
Eric Zelikman, Yuhuai Wu, Jesse Mu, and Noah Goodman.
\newblock Star: Bootstrapping reasoning with reasoning.
\newblock In \emph{Advances in Neural Information Processing Systems},
  volume~35, pages 15476--15488, 2022.

\bibitem[Zhao et~al.(2023)Zhao, Zhou, Li, Tang, Wang, Hou, Min, Zhang, Zhang,
  Dong, et~al.]{zhao2023survey}
Wayne~Xin Zhao, Kun Zhou, Junyi Li, Tianyi Tang, Xiaolei Wang, Yupeng Hou,
  Yingqian Min, Beichen Zhang, Junjie Zhang, Zican Dong, et~al.
\newblock A survey of large language models.
\newblock \emph{arXiv preprint arXiv:2303.18223}, 1\penalty0 (2):\penalty0
  1--124, 2023.

\end{thebibliography}

\end{document}